\documentclass[a4paper,fleqn]{cas-dc}

\usepackage[numbers]{natbib}

\def\tsc#1{\csdef{#1}{\textsc{\lowercase{#1}}\xspace}}
\tsc{WGM}
\tsc{QE}
\tsc{EP}
\tsc{PMS}
\tsc{BEC}
\tsc{DE}
\usepackage{tikz}
\usetikzlibrary{positioning}
\usepackage{pdflscape} 
\usepackage{rotating}
\usepackage{booktabs}  
\usepackage{siunitx}  
\usepackage{graphicx}
\usepackage{tikz}
\usepackage{adjustbox}
\usepackage{booktabs}
\usepackage{tikz}
\usetikzlibrary{mindmap, trees}
\usepackage{amssymb}
\usepackage{pifont}
\newcommand{\cmark}{\ding{51}}%
\newcommand{\xmark}{\ding{55}}%
\usepackage{tikz}
\usetikzlibrary{shadows}

\usetikzlibrary{mindmap, backgrounds}

\begin{document}
\let\WriteBookmarks\relax
\def\floatpagepagefraction{1}
\def\textpagefraction{.001}
\shorttitle{Task-based Loss Functions in Computer Vision: A Comprehensive Review}

\title [mode = title]{Task-based Loss Functions in Computer Vision: A Comprehensive Review} 

\author[1]{Omar Elharrouss}[]
\cormark[1]

\author[1]{Yasir Mahmood}[]
\cormark[1]
\author[1]{Yassine Bechqito}[]

\author[2]{Mohamed Adel Serhani}[]
\author[1]{Elarbi Badidi}
\author[3]{Jamal Riffi}[]

\author[3]{Hamid Tairi}[]

\address[1]{Department of Computer Science and Software Engineering, College of Information Technology, United Arab Emirates University.}
\address[2]{Department of Information Systems, College of Computing and Informatics, University of Sharjah, Sharjah, United Arab Emirates}
\address[3]{Department of Informatics, Faculty of Sciences Dhar El Mahraz, Sidi Mohamed Ben Abdellah University, Fez, Morocco}

\cortext[cor1]{Corresponding author}

\begin{abstract}
Loss functions are at the heart of deep learning, shaping how models learn and perform across diverse tasks. They are used to quantify the difference between predicted outputs and ground truth labels, guiding the optimization process to minimize errors. Selecting the right loss function is critical, as it directly impacts model convergence, generalization, and overall performance across various applications, from computer vision to time series forecasting. This paper presents a comprehensive review of loss functions, covering fundamental metrics like Mean Squared Error and Cross-Entropy to advanced functions such as Adversarial and Diffusion losses. We explore their mathematical foundations, impact on model training, and strategic selection for various applications, including computer vision (Discriminative and generative), tabular data prediction, and time series forecasting. For each of these categories, we discuss the most used loss functions in the recent advancements of deep learning techniques. Also, this review explore the historical evolution, computational efficiency, and ongoing challenges in loss function design, underlining the need for more adaptive and robust solutions. Emphasis is placed on complex scenarios involving multi-modal data, class imbalances, and real-world constraints. Finally, we identify key future directions, advocating for loss functions that enhance interpretability, scalability, and generalization, leading to more effective and resilient deep learning models.
\end{abstract}





\maketitle
\section{Introduction}

The progress in deep learning has been fueled by advancements in both model architectures and optimization techniques \cite{ii01}. Early deep learning models, primarily based on neural networks, relied on simple loss functions. However, the increasing complexity of deep learning models including Convolutional Neural networks (CNNs) and Vision transformers (ViTs), and the expansion into diverse application domains, such as computer vision, have necessitated the development of more sophisticated and specialized loss functions \cite{ii02}. These specialized functions are crucial for guiding the learning process effectively, particularly when addressing challenges like class imbalance, handling noisy data, and optimizing for specific task objectives.
\begin{table}[t!]
\textbf{List of Acronyms}\\
\begin{tabular}{ll}
ML                  & Machine Learning      \\
DL                  & Deep Learning         \\
CNN                  & Convolutional Neural Network \\
DNN                  & Deep Neural Network \\
ANN                  & Artificial Neural Network \\
ViT                   & Vision Transformer           \\
GAN          &Generative Adversarial Networks  \\
RPN          &Region Proposal Network   \\
YOLO          &You Only Look Once   \\
SSD          &Single Shot MultiBox Detector   \\
R-CNN          &Region-based CNN   \\
TSM          & Temporal Shift Module   \\
MViT          & Multiscale Vision Transformers  \\
HAT          & Hierarchical Action Transformer  \\
FCN          & Fully Convolutional Network  \\
PSPNet          & Pyramid Scene Parsing Network  \\
ENet          & Efficient Neural Network  \\
HRNet          & High Resolution Network  \\
LSTM          & Long Short-Term Memory  \\
RNN           & Recurrent Neural Network \\

ATOM          &Accurate Tracking by Overlap Maximization\\
DiMP          &Discriminative Model Prediction  \\
CLIP         & Contrastive Language-Image Pretraining  \\
FFC          &Fourier Convolution   \\
CMT          &Continuous Mask-aware Transformer  \\
OLS          &Ordinary Least Squares  \\
MLE          &Maximum Likelihood Estimation  \\  
MSE          &Mean Squared Error  \\
L1           &Mean Absolute Error   \\
SVM          &Support Vector Machines  \\
VAE          &Variational Autoencoders  \\
IOU          & Intersection Over Union  \\
CIOU          &Complete Intersection Over Union   \\
EIOU          & Efficient IOU   \\
PMSE          &    Predicted Mean Squared Error  \\
BCE          & Binary Cross-Entropy   \\
WCE          & Weighted Cross-Entropy  \\
BalanCE          & Balanced Cross-Entropy   \\
sDice          & Squared Dice  \\
lcDice          &  Log-Cosh Dice  \\
SSIM         &Structural Similarity Index Measure \\
MS-SSIM         &Multiscale Structural Similarity \\
\end{tabular}
\end{table}

Loss functions in deep learning play a critical role in optimizing neural networks during training by measuring the discrepancy between predicted outputs and actual ground truth labels \cite{ii1}. These functions are categorized into various types based on the task used. For example, for regression tasks involving continuous data, popular loss functions include Mean Squared Error (MSE) and Mean Absolute Error (MAE), which quantify the difference between predicted values and true labels. On the other hand, for classification tasks dealing with categorical data, loss functions such as Binary Cross-Entropy and Categorical Cross-Entropy are commonly employed to calculate the error in predicted class probabilities \cite{ii2}.

In addition to standard loss functions, specialized loss functions like Huber Loss, Hinge Loss, and Dice Loss are utilized for specific tasks that require different forms of error measurement \cite{ii3}. These custom loss functions cater to the unique requirements of tasks such as object detection, image segmentation, and natural language processing, where the standard loss functions may not be directly applicable. By selecting an appropriate loss function, deep learning models can effectively learn from data and optimize their parameters to improve predictive performance \cite{ii4}.


The choice of a suitable loss function in deep learning models has a significant impact on the training process and ultimately the model's accuracy  \cite{ii5,ii6}. A well-chosen loss function guides the optimization algorithm in the right direction, helping the model converge to a solution that minimizes the prediction error. Different loss functions can make learning from different aspects of the training process, such as penalizing outliers more harshly or handling class imbalances effectively \cite{ii7}. Therefore, understanding the characteristics and implications of various loss functions is essential for designing robust deep learning models that can generalize well to unseen data and perform effectively in real-world applications. Recently, in the same method, different loss functions can be fused to be used in training processes to cover different aspects, like image generation methods which need a set of limitations to be respected like the image resolution and the refinement.

This paper focuses on the pivotal role of loss functions in deep learning, particularly within the realm of computer vision and related areas.  In computer vision, we examine loss functions used in both discriminative and generative tasks.  Discriminative tasks, such as image classification, object detection, and semantic segmentation, rely heavily on loss functions to accurately measure the discrepancy between predicted labels and ground truth. Generative tasks, including text-to-image, image-to-image, and audio-to-image generation, use loss functions to evaluate the realism and quality of generated outputs, often using adversarial or perceptual losses to guide the training process.

Beyond computer vision, we also explore the application of loss functions in other data modalities, specifically tabular data prediction and time series forecasting. In tabular data prediction, appropriate loss functions are essential for guiding model learning on structured datasets, optimizing for prediction accuracy, and handling potential issues with missing or imbalanced data. Similarly, time series forecasting, which deals with sequential data exhibiting temporal dependencies, necessitates loss functions that can effectively capture these dependencies and produce accurate forecasts.

Also, this review paper provides a comprehensive overview of loss functions used across various deep learning tasks, emphasizing their historical development, current applications, and future research directions. We present a systematic categorization of loss functions by task type, describe their properties and functionalities, and analyze their computational implications.  By examining both theoretical and practical aspects, this paper aims to provide a valuable resource for researchers and practitioners seeking to improve the effectiveness of their deep learning models through careful selection and optimization of loss functions.

The content of this review is presented as follows: 
\begin{itemize}

    \item Summarization of the existing loss functions per task. 
    \item A description of different loss functions.
    \item A classification of the types of loss functions used in deep learning methods is performed.
    \item A discussion of computation evaluation of loss functions.
    \item Benefits, limitations, challenges, and future directions.

\end{itemize}

\begin{figure*}[t!]
\centering
  \begin{tabular}[b]{c}
    \includegraphics[width=.81\linewidth]{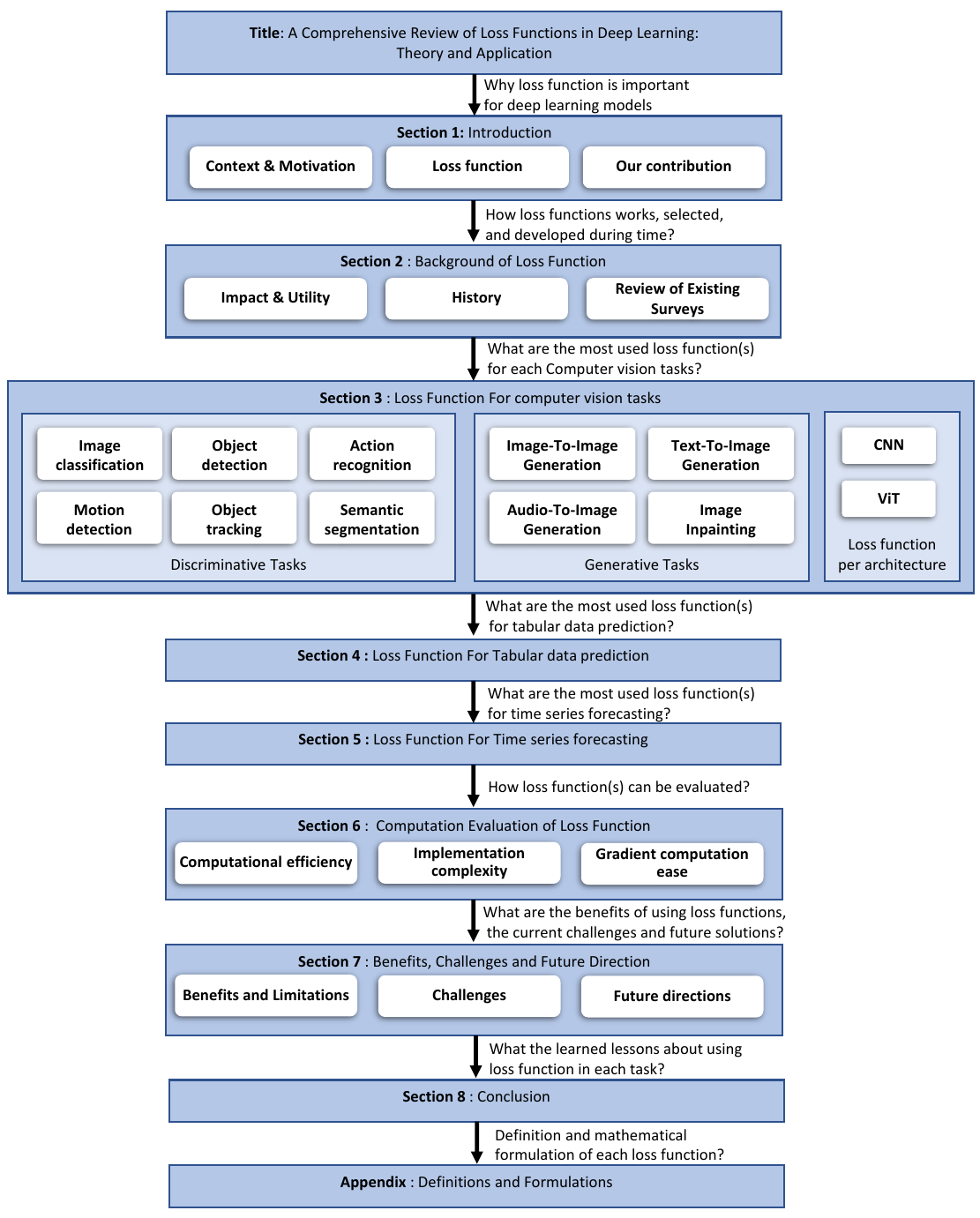}\\
  \end{tabular}
  \caption{{The taxonomy of the paper.}}
\label{paperorg}
\end{figure*}

The organization of the paper as illustrated in Figure \ref{paperorg} started with an introduction to the context of loss functions as well as their utilization in different deep learning models. The impact of loss functions on the development of artificial intelligence applications is provided in section 2 with a historical overview. In section 3, the loss functions used in various computer vision tasks are discussed with a presentation of the most used loss function in each task. Also in the same section, we compare the loss functions based on the architecture used including CNN-based models and ViT-based models. The methodology is presented in sections 4 and 5 for the used loss functions for the prediction from tabular data, and time series forecasting. A discussion of the computational evaluation of loss functions is presented in section 6. While, in section 7, we presented the advantages and disadvantages of some selected loss functions, challenges, and future directions. The conclusion is provided in section 8. In order to present all the cited loss functions, the definitions, and the mathematical formulations are provided in the Appendix.

\section{Background of Loss function}

The loss function is essential in training machine/deep learning models for measuring the difference between predictions and target results, and providing a clear metric for performance evaluation \cite{ii7}. It helps back-propagation to use the gradient of the loss function to adjust the parameters, also minimizing loss and improving accuracy. In addition, effective loss functions help balance bias and variance to avoid overfitting. 

\begin{figure*}[t!]
\centering
  \begin{tabular}[b]{c}
    \includegraphics[width=1\linewidth]{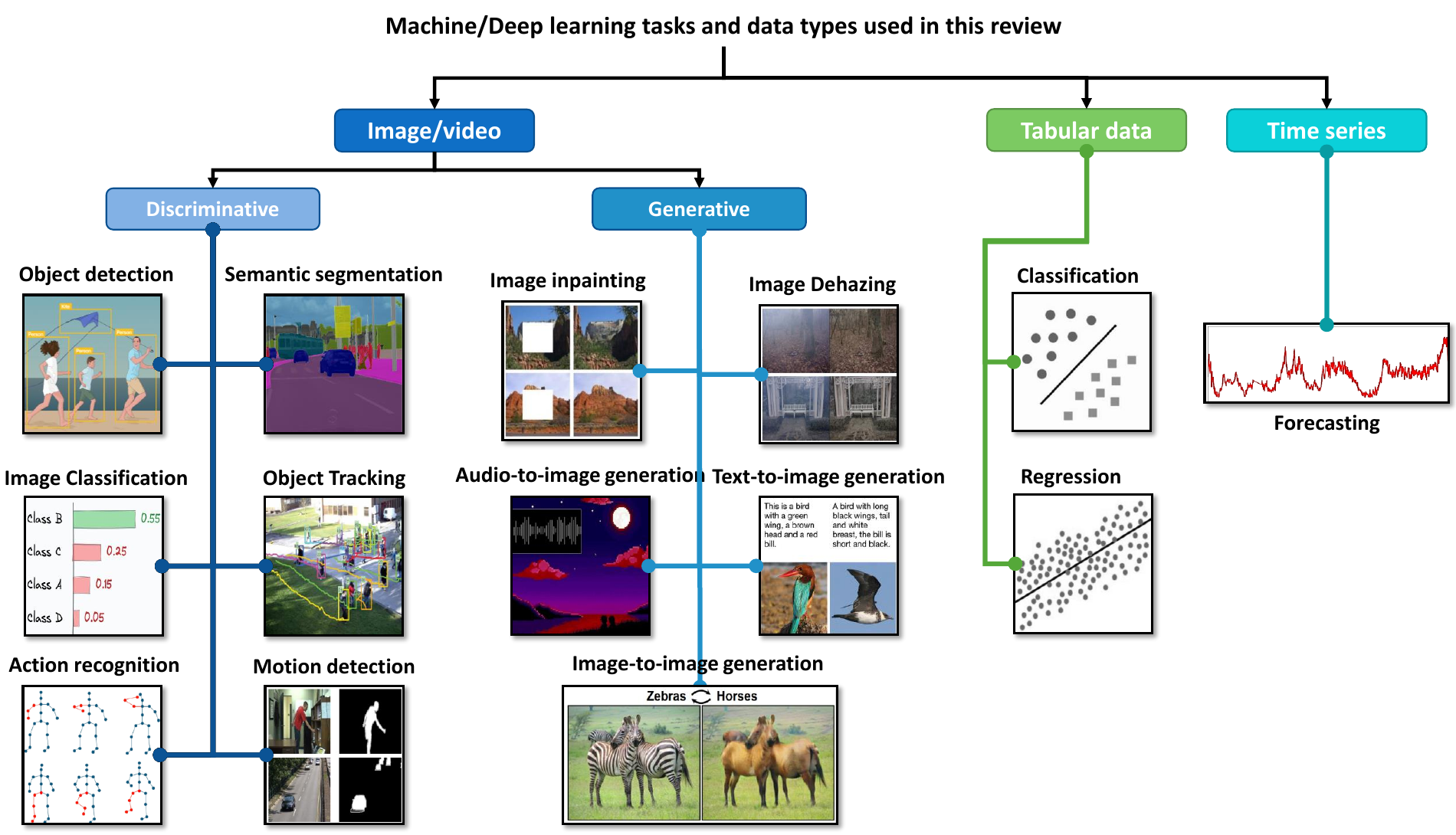}\\
  \end{tabular}
  \caption{{Various machine and deep learning tasks using different types of data that necessitate loss function for learning process.}}
\label{types}
\end{figure*}

Loss functions are categorized based on the addressed tasks, like regression and classification for tabular data, and image classification, detection, recognition, or generation for image processing presented in Figure \ref{types}. For regression tasks, models predict continuous output values, while for classification tasks, they produce discrete labels corresponding to dataset classes. The choice of loss function depends on the type of problem, and standard loss functions are typically aligned with either regression or classification tasks, to ensure effective model training and accurate predictions.

For computer vision tasks, the difference between loss functions for discriminative (like image classification) or generative (like image generation from different inputs) tasks centers on their objectives and evaluation methods. For example, image classification aims to categorize images into predefined classes using loss functions, which penalizes incorrect predictions to enhance accuracy. In contrast, image generation seeks to create new images similar to training data using various loss functions. For example, Generative Adversarial Networks (GANs) use adversarial loss to train a generator to produce realistic images that fool a discriminator. While, autoencoders employ reconstruction loss to ensure that generated images closely resemble the original inputs. Thus, classification loss functions prioritize prediction accuracy, while generative loss functions focus on creating outputs that mirror real data.

\subsection{Loss Function: Impact, Utilities, and History }

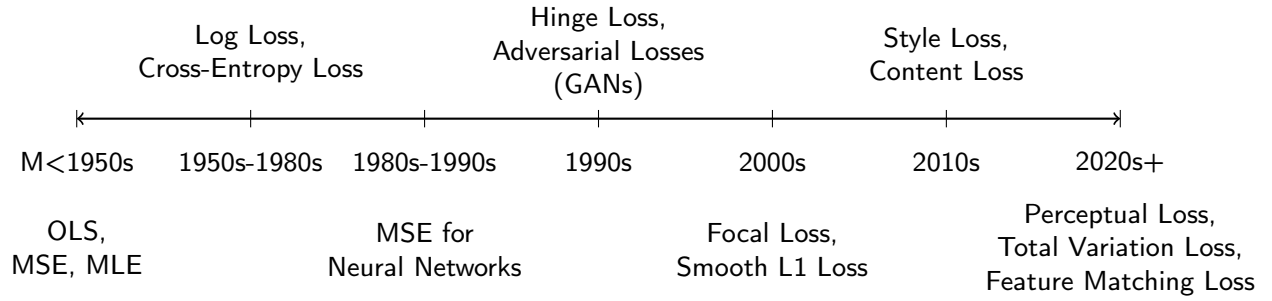
\begin{figure*}
\centering
\begin{tikzpicture}[scale=1.15, transform shape]
    \draw[<->, thick] (2, 0) -- (14, 0);

    \foreach \x in {2, 4, 6, 8, 10, 12, 14}
        \draw (\x, 0.1) -- (\x, -0.1);

    
    \node at (2, -0.5) {M<1950s};
    \node at (4, -0.5) {1950s-1980s};
    \node at (6, -0.5) {1980s-1990s};
    \node at (8, -0.5) {1990s};
    \node at (10, -0.5) {2000s};
    \node at (12, -0.5) {2010s};
    \node at (14, -0.5) {2020s+};

    \node[align=center, text width=3cm] at (2, -1.5) {OLS, \\ MSE, MLE};
    \node[align=center, text width=3cm] at (4, 0.75) {Log Loss, \\ Cross-Entropy Loss};
    \node[align=center, text width=3cm] at (6, -1.5) {MSE for \\ Neural Networks};
    \node[align=center, text width=3cm] at (8, 0.75) {Hinge Loss, \\ Adversarial Losses \\ (GANs)};
    \node[align=center, text width=3cm] at (10, -1.5) {Focal Loss, \\ Smooth L1 Loss};
    \node[align=center, text width=3cm] at (12, 0.75) {Style Loss, \\ Content Loss};
    \node[align=center, text width=4cm] at (14, -1.5) {Perceptual Loss, \\ Total Variation Loss, \\ Feature Matching Loss};
\end{tikzpicture}
    \caption{Timeline of the evolution of loss functions.}
    \label{timeline}
\end{figure*}

The evolution of loss functions reflects the transformative advancements in statistical methods and machine learning over the past century. Loss functions serve as critical components in model training, providing quantitative measures of how well a model's predictions align with actual outcomes. A historical description can highlight the transition from basic statistical techniques to sophisticated deep learning paradigms, illustrating the increasing complexity and specificity of loss functions designed to tackle diverse challenges across various domains. By understanding the development of loss functions, we gain insight into the foundational principles that drive modern machine learning and the methodologies that support effective predictive modeling. For that, the evolution of loss functions during time is presented in this section and illustrated in a timeline in Figure \ref{timeline}. 

\textbf{Statistical Methods (19th-20th Century)}: The concept of loss functions in statistical methods can be traced back to Ordinary Least Squares (OLS) \cite{old1}, used in linear regression. OLS aims to minimize the sum of squared differences between observed values and predicted values \cite{old2}. This foundational approach serves as an early example of using mathematical functions to minimize error and make predictions based on data.

\textbf{Early Machine Learning (Mid-20th Century):} As machine learning began to develop, Maximum Likelihood Estimation (MLE) became prominent \cite{mleold1}. MLE is a statistical method used to find parameter values that maximize the likelihood of the observed data fitting a particular model \cite{mleold2}. This often involves using loss functions like negative log-likelihood to optimize statistical models. We can find also Mean Squared Error (MSE) which is a standard criterion in statistical modeling and machine learning to evaluate the performance of forecasting and predictive models as these fields evolved throughout the 20th century \cite{mseold1}. MSE is sensitive to outliers because it squares the errors, meaning larger errors have a disproportionately high effect on the MSE value \cite{mseold2}.
\textbf{Perceptrons and Linear Models (1950s-1980s):} The perceptron, an early type of neural network, used a simple loss aimed at minimizing classification errors. In parallel, logistic regression introduced log loss \cite{oldlog1} (cross-entropy loss), which measures the discrepancy between actual labels and predicted probabilities. This period marks a shift toward optimizing functions to improve classification accuracy.

\textbf{Artificial Neural Networks (1980s-1990s):} With the development of neural networks, Mean Squared Error (MSE) became commonly used for regression tasks. MSE calculates the average squared differences between predicted and actual values, guiding adjustments in model training to minimize errors and enhance prediction precision.

\textbf{Support Vector Machines and Deep Learning (1990s-Present):} The introduction of Support Vector Machines (SVMs) brought hinge loss \cite{hingeloss}, which maximizes the margin between classes for classification tasks. In deep learning, cross-entropy loss grew in popularity, effectively handling multi-class classification by measuring dissimilarity between predicted probabilities and actual classes.

\textbf{Modern Advances (2010s-Present):} In recent years, new paradigms like Generative Adversarial Networks (GANs) have emerged, using adversarial losses \cite{adversarial} to train models in competitive settings. Variational Autoencoders (VAEs) combine reconstruction loss with Kullback-Leibler divergence for learning latent representations. Reinforcement learning utilizes reward signals as a form of loss to optimize decision-making policies, illustrating the diverse applications of loss functions in contemporary machine learning.

For training the deep learning models loss functions are fundamental which can make the learning very effective. Their impacts can be on different parts including parameter updates and objective measurement. For example, a loss function provides a direction for updating the model's parameters (weights and biases) during training. The gradient of the loss function with respect to these parameters indicates how much to change to reduce the loss. Also, it can be used as a metric to evaluate how well a model is performing, which is essential for comparing models and adjusting training processes. While,
the choice of loss function can significantly affect the model's performance. A well-chosen loss function can lead to faster convergence and better prediction accuracy.

\begin{table*}[t!]
\scriptsize
\center
\caption{Existing surveys on loss functions compared with the proposed review.}
\label{table:review}
\begin{tabular}{|p{1.7cm}|p{.7cm}|c|c|c|c|c|c|c|c|c|c|c|p{6cm}|}
\hline
 \multirow{11}{*}{\textbf{Reference}} & \multirow{11}{*}{\rotatebox{90}{\textbf{Jour/Conf}}}&\multicolumn{2}{|c|}{\textbf{ML}}&
  \multicolumn{9}{|c|}{\textbf{DL}} & \multirow{11}{*}{\textbf{Description}} \\\cline{3-13}

&&  \rotatebox{90}{\textbf{Classification}}  &\rotatebox{90}{\textbf{Regression}} & \rotatebox{90}{\textbf{Image classification  }} & \rotatebox{90}{\textbf{Object detection }} & \rotatebox{90}{\textbf{Object tracking }} & \rotatebox{90}{\textbf{Action recognition }} & \rotatebox{90}{\textbf{Semantic segment }}&
\rotatebox{90}{\textbf{Image-to-image gen  }} & \rotatebox{90}{\textbf{Text-to-image gen }} & \rotatebox{90}{\textbf{Audio-to-image gen }} & \rotatebox{90}{\textbf{Image Inpainting }}& \\ \cline{1-14}


Jadon et al. (2020) \cite{r1} &  \multirow{5}{*}{Conf}&  \multirow{5}{*}{\xmark} & \multirow{5}{*}{\xmark} & \multirow{5}{*}{\xmark} & \multirow{5}{*}{\xmark} & \multirow{5}{*}{\xmark} & \multirow{5}{*}{\xmark} & \multirow{5}{*}{\cmark} & \multirow{5}{*}{\xmark} & \multirow{5}{*}{\xmark} & \multirow{5}{*}{\xmark} & \multirow{5}{*}{\xmark} & 
Reviewed recent advancements in image segmentation using various objective loss functions introduces a new log-cosh dice loss function, comparing its performance on the NBFS skull-segmentation dataset with other common functions.\\
\hline

Jurdi et al. (2021) \cite{r2} & \multirow{5}{*}{Journ}  &  \multirow{5}{*}{\xmark} & \multirow{5}{*}{\xmark} & \multirow{5}{*}{\xmark} & \multirow{5}{*}{\xmark} & \multirow{5}{*}{\xmark} & \multirow{5}{*}{\xmark} & \multirow{5}{*}{\cmark} & \multirow{5}{*}{\xmark} & \multirow{5}{*}{\xmark} & \multirow{5}{*}{\xmark} & \multirow{5}{*}{\xmark} & Discussed the use of deep convolutional neural networks (CNNs) in medical image segmentation. It focuses on integrating high-level prior knowledge into loss functions to enhance segmentation.  \\
\hline

Wang et al. (2020) \cite{r3} & \multirow{3}{*}{Journ} &   \multirow{3}{*}{\cmark} & \multirow{3}{*}{\cmark} & \multirow{3}{*}{\xmark} & \multirow{3}{*}{\cmark} & \multirow{3}{*}{\xmark} & \multirow{3}{*}{\xmark} & \multirow{3}{*}{\xmark} & \multirow{3}{*}{\xmark} & \multirow{3}{*}{\xmark} & \multirow{3}{*}{\xmark} & \multirow{3}{*}{\xmark} & 
Highlighted the used loss functions in machine learning and deep learning. It reviews 31 loss functions, categorizing them by traditional ML tasks and DL applications. \\ \hline

Tian et al. (2022) \cite{r4} &\multirow{5}{*}{Journ}  &   \multirow{5}{*}{\xmark} & \multirow{5}{*}{\xmark} & \multirow{5}{*}{\xmark} & \multirow{5}{*}{\cmark} & \multirow{5}{*}{\xmark} & \multirow{5}{*}{\xmark} & \multirow{5}{*}{\cmark} & \multirow{5}{*}{\xmark} & \multirow{5}{*}{\xmark} & \multirow{5}{*}{\xmark} & \multirow{5}{*}{\xmark} &
Discusses the importance of loss functions in training neural networks and their limitations in deep learning. It reviews recent advancements in loss functions for some CV tasks.  \\\hline

Azad et al. (2023) \cite{r5} & \multirow{3}{*}{Preprint} &     \multirow{3}{*}{\xmark} & \multirow{3}{*}{\xmark} & \multirow{3}{*}{\xmark} & \multirow{3}{*}{\xmark} & \multirow{3}{*}{\xmark} & \multirow{3}{*}{\xmark} & \multirow{3}{*}{\cmark} & \multirow{3}{*}{\xmark} & \multirow{3}{*}{\xmark} & \multirow{3}{*}{\xmark} & \multirow{3}{*}{\xmark} &
Provides a comprehensive review and novel taxonomy of loss functions used to enhance semantic image segmentation, categorizing them by features and applications.\\ 
\hline

Terven et al. (2023) \cite{r6} & \multirow{4}{*}{Preprint}&    \multirow{4}{*}{\xmark} & \multirow{4}{*}{\xmark} & \multirow{4}{*}{\xmark} & \multirow{4}{*}{\xmark} & \multirow{4}{*}{\xmark} & \multirow{4}{*}{\xmark} & \multirow{4}{*}{\xmark} & \multirow{4}{*}{\xmark} & \multirow{4}{*}{\xmark} & \multirow{4}{*}{\xmark} & \multirow{4}{*}{\xmark} 
&Reviewed the importance of loss functions and performance metrics in deep learning. It reviews common loss functions and measurements, analyzing their benefits and limitations across various DL problems.
 \\ \hline
Ciam et al. (2023) \cite{r7} & \multirow{3}{*}{Preprint}  &   \multirow{3}{*}{\cmark} & \multirow{3}{*}{\cmark} & \multirow{3}{*}{\xmark} & \multirow{3}{*}{\xmark} & \multirow{3}{*}{\xmark} & \multirow{3}{*}{\xmark} & \multirow{3}{*}{\xmark} & \multirow{3}{*}{\xmark} & \multirow{3}{*}{\xmark} & \multirow{3}{*}{\xmark} & \multirow{3}{*}{\xmark} &
 Highlights the critical role of loss functions in optimizing machine learning techniques. It offers an overview of 43 widely used loss functions across key ML applications. \\ \hline

\multirow{8}{*}{\textbf{Proposed}} & \multirow{8}{*}{\textbf{-}}  &   \multirow{8}{*}{\cmark} & \multirow{8}{*}{\cmark} & \multirow{8}{*}{\cmark} & \multirow{8}{*}{\cmark} & \multirow{8}{*}{\cmark} & \multirow{8}{*}{\cmark} & \multirow{8}{*}{\cmark} & \multirow{8}{*}{\cmark} & \multirow{8}{*}{\cmark} & \multirow{8}{*}{\cmark} & \multirow{8}{*}{\cmark} & We explored common loss functions in machine and deep learning for computer vision, tabular data prediction, and time series forecasting. We examined works using techniques like CNNs, ViTs, and diffusion models, often combining loss functions for image generation. The tasks include classification, regression, and various computer vision applications such as discriminative and generative tasks.
 \\ \hline

\end{tabular}

\end{table*}

\subsection{Previous reviews and surveys}

In this section, reviews and surveys on loss functions for various purposes were examined \cite{r1,r2,r3,r4,r5,r6,r7}. 
These reviews can be categorized based on the task used. For example, reviews on image segmentation are discussed in \cite{r1, r2}, and semantic segmentation in \cite{r5}. Some of them are based on the techniques used such as machine learning like in \cite{r3,r7}, and deep learning like in \cite{r6}. The existing works review key advancements and challenges related to loss functions, particularly focusing on their applications in image segmentation and other computer vision tasks. They emphasize the importance of selecting appropriate loss functions to enhance model performance across various tasks such as object detection, face recognition, and segmentation.

Several papers propose novel loss functions to address specific issues like imbalanced data and anatomical accuracy. Comprehensive reviews categorize existing loss functions, analyze their properties, and discuss their applications, aiming to assist researchers and practitioners in choosing the optimal methods for their tasks. Some works also highlight future research directions and challenges in the field, such as designing robust losses and improving evaluation metrics. Overall, these surveys serve as valuable resources for students and researchers looking to deepen their understanding of loss functions in machine learning.
For example, the authors in \cite{r1} summarized various objective loss functions for image segmentation. Also identified certain loss functions that perform well across multiple datasets. In the same context but for medical image segmentation, the authors in \cite{r2}  focused on integrating high-level prior knowledge into loss functions to enhance segmentation. For an advanced segmentation task which is semantic segmentation, the authors in \cite{r5}  highlighted the role of loss functions in enhancing semantic segmentation algorithms. It provides a comprehensive review and novel taxonomy of loss functions used in image segmentation, categorizing them by features and applications.
While, in \cite{r4} the authors reviewed recent advancements in loss functions for some computer vision tasks including object detection, face recognition, and image segmentation. Also discussed the importance of loss functions in training neural networks and their limitations in deep learning. 

For Machine and deep learning techniques, the authors in \cite{r3} Highlighted the importance of loss functions in machine learning and the need for a comprehensive analysis of classical loss functions. It reviews 31 loss functions, categorizing them by traditional machine learning tasks (classification, regression, unsupervised learning) and deep learning applications (object detection, face recognition). 
While the authors in \cite{r6} reviewed loss functions and performance metrics in deep learning algorithms. It reviews common loss functions and measurements, analyzing their benefits and limitations across various deep-learning problems. 
In \cite{r7} the authors highlight the critical role of loss functions in optimizing machine learning techniques. It offers an overview of 43 widely used loss functions across key applications like regression, classification, and generative modeling, organized within a clear taxonomy that outlines their theoretical foundations and optimal contexts.

In our proposed review, we examined the most commonly used loss functions applied in machine and deep learning techniques for computer vision, tabular data prediction, and time series forecasting. For each task, we collected advanced works that utilized different deep learning techniques, including CNNs, Vision Transformers (ViTs), and diffusion models, along with various loss functions. Some of these methods employed a combination of more than one loss function, especially for image generation models. In the reviewed tasks, we identified machine learning tasks, including classification and regression, as well as computer vision tasks such as image classification, object detection, semantic segmentation, action recognition, object tracking, and image generation. 

To compare the existing surveys with the proposed review, Table \ref{table:review} summarizes each review in terms of tasks and loss function used.


\section{Loss function for computer vision tasks}


Computer vision tasks using deep learning require distinct approaches to measuring error, as the nature of the data and the objectives can vary significantly. These tasks can be categorized into two main categories: discriminative and generative tasks. While discriminative tasks can include classification, detection, recognition and segmentation. For instance, image classification focuses on determining the correct category for an image, while object detection involves not only classification but also the identification of object locations within an image. Similarly, semantic segmentation allocates a label to every pixel in an image, requiring a more complex loss structure to encourage spatial coherence.

Generative tasks include all image generation using different input data like text, audio, and image.
For example, image generation relies on adversarial frameworks, where loss functions reflect the balance between generated and real images, while tabular data prediction and time series forecasting may require regression-oriented loss functions to handle continuous outputs.

In this section, we discuss the various loss functions designed for distinct deep learning tasks, highlighting their suitability to specific requirements. By understanding these functions, researchers can make informed decisions about model selection, tuning, and improvements, ultimately enhancing the efficacy of their predictive frameworks.

\subsection{Discriminative tasks}
Discriminative tasks in computer vision focus on the analysis and interpretation of visual data to make informed classifications and identifications. Unlike generative tasks, which are concerned with creating new images or transforming existing ones, discriminative tasks aim to differentiate and categorize images based on their content. This includes image classification, object detection, action recognition, and semantic segmentation.

The core of discriminative tasks is the ability to identify specific features, patterns, and relationships within images. For example, image classification assigns a label to an entire image, enabling systems to categorize visual data efficiently. While, Object detection takes this a step further by not only identifying objects but also locating them within the image using bounding boxes.

Advanced machine learning techniques, particularly convolutional neural networks (CNNs) and other deep learning models have significantly enhanced the accuracy and efficiency of these tasks. Loss functions help to improve these tasks by measuring the prediction or generation error during the learning process. By the following a revision of various loss functions used in each task.

\begin{figure*}[t!]
\centering
\includegraphics[width=.71\linewidth]{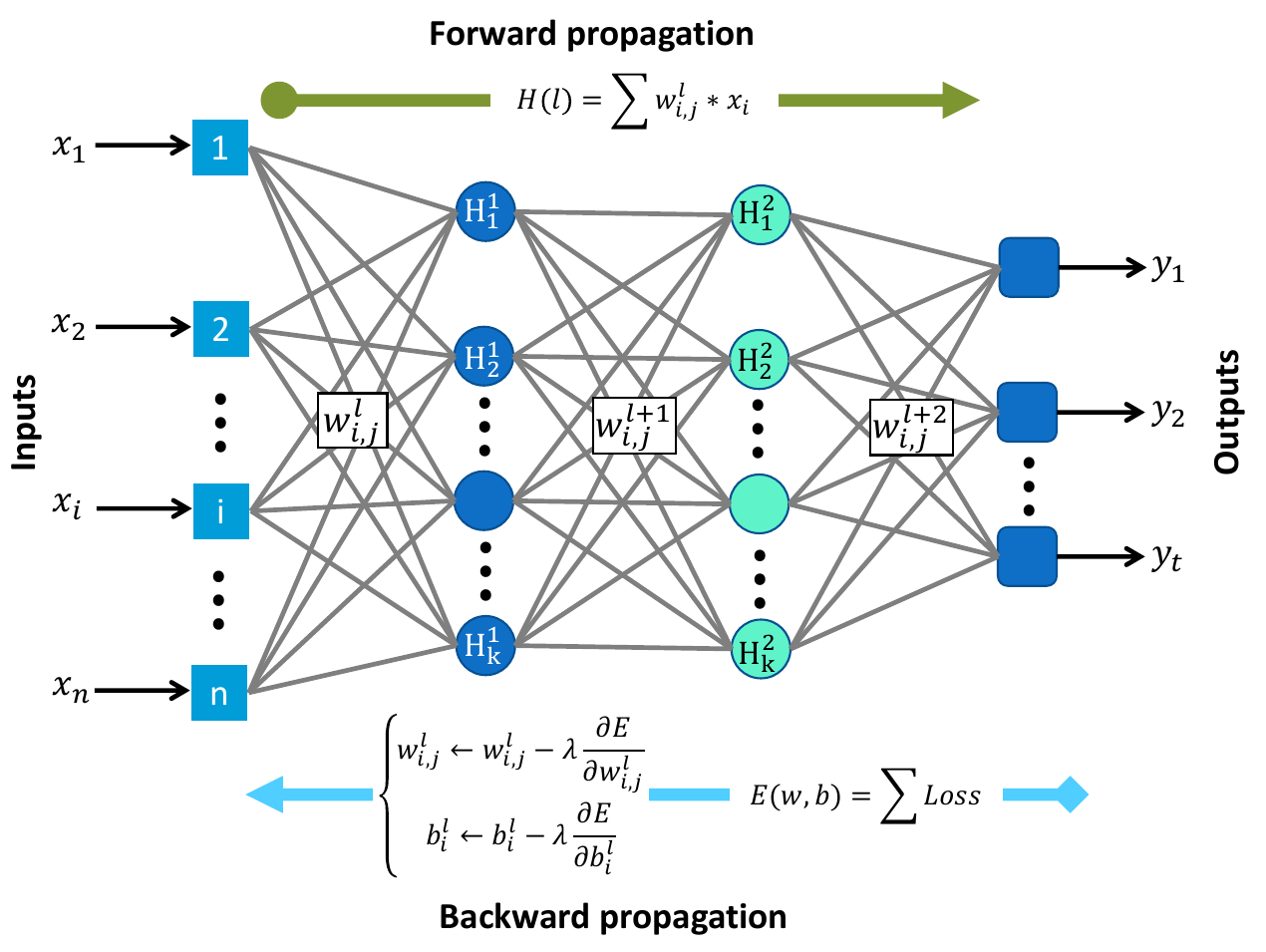}
\caption{Neural network forward and backward propagation.}
\label{class}
\end{figure*}

\textbf{Image classification:}
Loss functions in image classification, particularly in deep learning, play a crucial role in evaluating how well a model's predictions match the true labels of the images by utilizing backpropagation to compute the gradients of the loss concerning the model parameters presented in Figure \ref{class}. Common loss functions used include Cross-Entropy Loss and Mean Squared Error (MSE). Cross-entropy loss is widely used for classification tasks because it measures the performance of a classification model whose output is a probability value between 0 and 1, which means minimizing wrong classifications more. It tends to encourage the model to output higher probabilities for correct classes, improving accuracy in multi-class problems.  While, Mean Squared Error (MSE), though more common in regression, can be applied to image classification but is generally less effective as it doesn't handle probabilities as efficiently as Cross-Entropy Loss. The choice and effectiveness of the loss function significantly impact the model's ability to learn and generalize from the training data, leading to better performance on unseen images.

From deep learning models that are widely used for image classification can be taken into consideration along with the related loss functions used in training these models. From these models we can find the CNN-based models that are used as a backbone for many computer vision tasks and trained for image classification including VGG \cite{VGG}, Inception \cite{Inceptionv3},  WideResNet \cite{WideResNet}. While VGG used Euclidean loss, Inception and WideResNet used cross-entropy as loss function. 

In the same context, using the Vision Transformer (ViT) networks cross-entropy is used also to train these methods. For example, in \cite{RNAViT} which is a model that improves private inference efficiency for ViTs by leveraging reduced-dimension attention and novel softmax approximation techniques, used cross-entropy as a loss function. While, in \cite{TrojViT} Troj-ViT which is a ViT-based network designed for backdoor attacks on ViTs, this method uses patch-wise triggers to manipulate ViT parameters subtly, maintaining normal inference accuracy while enabling targeted class miss-classification upon trigger activation. TrojViT merged 
cross-entropy loss and Attention-Target loss (defined in \cite{TrojViT}) for training the network. Another loss function named Ranking loss \cite{Ranking} is utilized to maintain the relative order of attention scores before and after quantization. This loss function is used in \cite{RepQViT} RepQ-ViT that addresses post-training quantization challenges in ViTs. This method decouples quantization from inference, using complex quantizers for accurate quantization and simplified ones for efficient inference, improving accuracy in low-bit scenarios.
Cross-entropy and  Kullback-Leibler divergence loss functions are used in MGViT \cite{MGViT} and LFViT \cite{LFViT}. While, LF-ViT \cite{LFViT} reduces computational requirements by processing low-resolution images initially and focusing on specific regions, achieving notable reductions in FLOPs and improvements in throughput.

\textbf{Object Detection:}
Loss functions for object detection in deep learning are pivotal in evaluating how well a model can identify and locate objects within an image. Commonly employed loss functions include Region-based loss functions like the Region Proposal Network (RPN), Classification Loss, and Localization Loss (or Bounding Box Regression Loss). The Classification Loss or Cross-Entropy Loss, is used to determine how well the model can classify each proposed region as an object of interest or background. This encourages the model to correctly identify and differentiate between various classes. From the known methods that used Cross-Entropy loss we can find FastRCNN \cite{fastrcnn} and FasterRCNN \cite{fasterrcnn}.

In the context of object detection, Localization Loss represented as Smooth L2 Loss or Mean Squared Error (MSE), measures the accuracy of predicted bounding box coordinates against the ground-truth coordinates. This loss helps in refining the bounding box predictions to more accurately capture the object's extent within the image. These loss functions are used in You Only Look Once (YOLO) methods with various versions \cite{yolo1,yolo2,yolo3,yolo4,yolo5,yolo6} as well as in Single Shot MultiBox Detector (SSD) method \cite{ssd}.

Gaussian Wasserstein Distance (GW Distance) loss function is used also for object detection to improve the way models evaluate the performance of predicted bounding boxes against ground-truth boxes. It leverages the principles of optimal transport and Wasserstein distance, providing a robust metric for comparing distributions, particularly in the context of Gaussian-distributed object proposals \cite{GW1}.
In object detection, bounding box predictions are often evaluated based on aspects like position, width, height, and sometimes the class \cite{GW2}. When employing Gaussian Wasserstein Distance Loss, the idea is to model the predicted bounding boxes and the ground-truth bounding boxes as Gaussian distributions.

The Efficient IOU loss function defined in \cite{EIOU} and used in \cite{EIOU1} separates the aspect ratio based on the Complete Intersection Over Union (CIOU) loss function. It calculates the difference between the width and height of the prediction box and the minimum outer rectangle, respectively, which can reflect the actual difference between the width and height, thus accelerating the convergence speed of the network. In addition, the EIOU loss function also adds the Focal idea; the Focal idea can deal with the uneven problem of sample classification, realize the detection of complex samples, and help improve the network’s training effect. Based on the Focal idea, Focal loss is another loss function used by EfficentDet \cite{efficientDet} and RetinaNet \cite{RetinaNet} which are widely used methods for object detection.

\begin{table*}[t!]
\centering
\footnotesize
\caption{Loss Functions used in various discriminative tasks. }
\label{descriminativeloss}
\begin{tabular}{|p{2cm}|p{3cm}|p{2cm}|p{6.5cm}|}
\hline
\textbf{Task} & \textbf{Method} & \textbf{Technique} &\textbf{Loss Function} \\
\hline \hline
\multirow{8}{*}{\textbf{Image }}&  VGG \cite{VGG} & CNN &Euclidean \\\cline{2-4}
 \multirow{9}{*}{\textbf{Classification}}& Inceptionv3 \cite{Inceptionv3}& CNN &Cross-Entropy \\ \cline{2-4}
& WideResNet\cite{WideResNet} & CNN &Cross-Entropy \\ \cline{2-4}
&  RNAViT \cite{RNAViT} & CNN &Cross-Entropy \\ \cline{2-4}
&  TrojViT \cite{TrojViT} & ViT &Cross-Entropy + Attention-Target\\\cline{2-4}
& RepQViT \cite{RepQViT} & ViT &Ranking \\ \cline{2-4}
& MGViT \cite{MGViT}  & ViT &Cross-Entropy + Kullback-Leibler  \\\cline{2-4}
&  LFViT \cite{LFViT} & ViT &Cross-Entropy + Kullback-Leibler  \\
\hline \hline

\multirow{8}{*}{\textbf{Object}}& FastR-CNN \cite{fastrcnn} & CNN &Cross-Entropy \\\cline{2-4}
\multirow{9}{*}{\textbf{ Detection}} & FasterR-CNN \cite{fasterrcnn}& CNN &Cross-Entropy, Smooth L1, Binary Cross-Entropy \\\cline{2-4}
&YOLOs \cite{yolo1,yolo6}& CNN &MSE, Smooth L1 \\\cline{2-4}
&YOLO2-3 \cite{yolo2,yolo3}& CNN &Intersection over Union (IoU), MSE, Smooth L1 \\\cline{2-4}

& SSD \cite{ssd}& CNN &MSE, Smooth L1, Focal, Binary Cross-Entropy \\\cline{2-4}
&Yang et al. \cite{GW1}& Statistical &Gaussian  Wasserstein distance \\\cline{2-4}
&Wang et al. \cite{GW2}& Statistical &Gaussian  Wasserstein distance \\\cline{2-4}
&Wang et al. \cite{EIOU,EIOU1}& Statistical & Focal, Efficient IOU  \\\cline{2-4}
& EfficentDet \cite{efficientDet}& CNN &Focal\\ \cline{2-4}
& RetinaNet \cite{RetinaNet}& CNN &Focal, Smooth L1 \\ \hline \hline

\multirow{8}{*}{\textbf{Action}}& Wang et al. \cite{wang2018nonlocal} & CNN &Cross-Entropy \\ \cline{2-4}
\multirow{9}{*}{\textbf{Recognition}}& TSM \cite{wu2019temporal} & CNN &Cross-Entropy \\ \cline{2-4}
& SlowFast \cite{feichtenhofer2019slowfast} & CNN &Cross-Entropy \\ \cline{2-4}
& MViT  \cite{fan2022mvit} & CNN, ViT &Cross-Entropy \\ \cline{2-4}
&TimeSformer \cite{bertasius2021timesformer}& ViT &Cross-Entropy \\ \cline{2-4}
&Video-Swin-Tr \cite{liu2021swin} & ViT &Cross-Entropy \\ \cline{2-4}
&HAT \cite{zhou2022hat} & ViT &Cross-Entropy \\ \cline{2-4}
& ActionCLIP \cite{lei2022actionclip}  & CNN, ViT &Contrastive \\ \hline\hline
\multirow{10}{*}{\textbf{Semantic}} & FCN \cite{fcn}& CNN &Pixel-wise Cross-Entropy, Weighted Cross-Entropy \\\cline{2-4}
\multirow{11}{*}{\textbf{Segmentation}}  & UNet \cite{unet}& CNN &Binary Cross-Entropy, Dice\\\cline{2-4}
& SegNet \cite{segnet}& CNN &Pixel-wise Softmax with Cross-Entropy\\\cline{2-4}
& DeepLab \cite{deeplab}& CNN &Pixel-wise Cross-Entropy, Focal\\\cline{2-4}
& PSPNet \cite{pspnet}& CNN &Pixel-wise Cross-Entropy, Focal\\\cline{2-4}
& ENet \cite{enet}& CNN &Pixel-wise Cross-Entropy or Focal\\\cline{2-4}
& BiFusion \cite{bifusion}& CNN &Pixel-wise Cross-Entropy,  Dice\\\cline{2-4}
& HRNet \cite{hrnet}& CNN &Pixel-wise Cross-Entropy or Focal\\\cline{2-4}
& Tri-Net \cite{trinet}& CNN &Cross-Entropy,  Dice\\ \cline{2-4}

& Attention-Unet \cite{attentionunet}& CNN, ViT&Binary Cross-Entropy or Dice\\\cline{2-4}
&SegFormer \cite{segformer}& ViT &Pixel-wise Cross-Entropy or Focal\\\cline{2-4}
& Segmenter\cite{segmenter} & ViT &Cross-Entropy\\\cline{2-4}
&Swin Transformer \cite{swinTR}& ViT &Cross-Entropy, Dice or Jaccard\\ \cline{2-4}
& DPT \cite{dpt}& ViT &Cross-Entropy, L1\\ \cline{2-4}
&MaskFormer \cite{maskformer}& ViT &Binary Cross-Entropy loss, Focal\\
\hline \hline

\multirow{5}{*}{\textbf{Motion}} &Deep-BS \cite{kim2019deep}&CNN & Binary Cross-Entropy  \\ \cline{2-4}
\multirow{6}{*}{\textbf{Detection}} &Goyal et al. \cite{goyal2020deep}&CNN  & Reconstruction Loss \\ \cline{2-4}
&Li et al. \cite{li2018motion}&CNN  & Cross-Entropy  \\\cline{2-4}
&Jha et al. \cite{jha2018deep}&LSTM  & MSE \\ \cline{2-4}
&Chen et al. \cite{chen2023spatiotemporal}&CNN &  Motion\\ \cline{2-4}
&BS-Optical-Flow \cite{wang2021motion}  &CNN &MSE, Motion Consistency \\ \hline\hline


\multirow{8}{*}{\textbf{Object}}  &STTA \cite{STTA} &CNN & Cross-Entropy, Triplet \\ \cline{2-4}
\multirow{9}{*}{\textbf{Tracking}} &SiamFC \cite{bertinetto2016fully} &ANN  & Cross-Correlation \\ \cline{2-4}
&Deep SORT \cite{wojke2017simple} &Kalman Filter  & Mahalanobis Distance \\ \cline{2-4}
&Track R-CNN \cite{voigtlaender2019track}&CNN  &  Softmax Cross Entropy, Smooth L1 \\ \cline{2-4}
&ATOM \cite{danelljan2019atom}&CNN, ResNet  & IoU \\ \cline{2-4}
&DiMP \cite{bhat2019learning}&CNN, ResNet  &  Logistic Regression \\ \cline{2-4}
&OSTrack \cite{xu2021ostra}&CNN  &  IoU, L1\\ \cline{2-4}
&TransTrack \cite{li2022transT}&ViT  & Focal, IoU, L1  \\ \cline{2-4}
&FairMOT \cite{wang2020fairmot}&CNN  &  Re-Identification \\ \cline{2-4}

&Diffusiontrack \cite{Diffusiontrack}&Diffusion  & Focal, GIoU, L1  \\ \cline{2-4}
&Motiontrack \cite{Motiontrack}& ViT  & L1\\ \hline
\end{tabular}
\end{table*}

\textbf{Action Recognition:}
Loss functions for action recognition and abnormal event detection in deep learning are integral to training models that accurately identify and interpret activities or detect unusual events within video data. For action recognition, where the task is to classify a sequence of frames into specific actions, Cross-Entropy loss is commonly employed. This loss function compares the predicted action class probabilities with the true action labels, helping to refine the model's ability to distinguish between different actions over time. Temporal loss functions, such as Temporal Cross-Entropy loss, can also be used to account for the sequential nature of video data, ensuring that the model captures temporal dependencies effectively.

Recent advancements in action recognition have introduced several innovative methods. For example, the Non-local Networks proposed by Wang et al. \cite{wang2018nonlocal} leverage a non-local operation to capture long-range dependencies in video sequences, using Cross-Entropy loss for training. The Temporal Shift Module (TSM) developed by Wu et al. \cite{wu2019temporal} allows for efficient modeling of temporal dynamics with minimal computational overhead, also utilizing Cross-Entropy loss. SlowFast Networks like in \cite{feichtenhofer2019slowfast}, which continue to evolve for improved action recognition performance, rely on Cross-Entropy Loss to develop the proposed method. Using ViT-based, MViT (Multiscale Vision Transformers) \cite{fan2022mvit} adapted vision transformers for multi-scale video data processing, employing Cross-Entropy loss. While, TimeSformer is a transformer architecture specifically designed for video understanding, using Cross-Entropy loss \cite{bertasius2021timesformer}. The Video Swin-Transformer adapted the Swin Transformer for video classification tasks and uses Cross-Entropy loss \cite{liu2021swin}. Using the same loss function, the Hierarchical Action Transformer (HAT) proposed by Zhou et al. \cite{zhou2022hat} captures hierarchical relationships through multiple temporal scales.
Another loss function is implemented in \cite{lei2022actionclip} and named Contrastive loss is used in ActionCLIP which is a contrastive learning technique for zero-shot action recognition.

\textbf{Semantic Segmentation:}
Image segmentation aims to analyze an image and assign a specific category to each pixel. The model should identify and classify the pixels that belong to each category which can be an animal, the grass, the sky, or other elements within the image. To measure the model’s performance, one or more loss functions can be utilized. These functions compare the predicted and the generated labels by evaluating how accurately the model is performing and guiding the learning process. 

In literature, many semantic segmentation surveys discussed the loss functions used in CNN-based \cite{segloss} or ViT-based \cite{segloss1} techniques. For example, the authors in \cite{segloss} split the loss functions used for semantic segmentation into three categories: Distribution-based loss functions, region-based loss functions, and Compound loss functions. While the distribution-based loss category is a measure of the distance between the predicted and true values in a pixel-by-pixel fashion. This category includes Binary Cross-Entropy (BCE), Weighted Cross-Entropy (WCE),and Balanced Cross-Entropy (BalanCE). The region-based loss type measures the non-overlap between the road segmentation map and the ground truth map and contains Jaccard loss, Dice loss, Squared Dice (sDice) loss, Log-Cosh Dice (lcDice) loss, and Tversky loss. On the other hand, the compound loss type is a combination of both types, thereby leveraging pixel- and region-level losses and composed of BCE-Dice loss and Combo loss. The loss functions and others are used in different deep-learning semantic segmentation methods. Some of these methods are widely used due to their effectiveness. From these methods we can find CNN-based methods like FCN \cite{fcn}, Unet \cite{unet}, SegNet \cite{segnet}, DeepLab \cite{deeplab}, PSPNet \cite{pspnet}, and ViT-based methods like Segmenter \cite{segmenter}, SegFormer \cite{segformer}, Swin-Transformer \cite{swinTR}, MaskFormer \cite{maskformer}. 

CNN-based methods use various loss functions or a combination of them to implement their methods. For example, Fully Convolutional Networks (FCN) \cite{fcn} are among the widely used architectures for semantic segmentation, commonly utilizing pixel-wise Cross-Entropy loss or Weighted Cross-Entropy loss to account for class imbalance in the dataset. The same for U-Net \cite{unet} which is widespread recognition in image segmentation, typically employing Binary Cross-Entropy loss, Dice loss, or combinations of both to enhance performance, particularly in cases with imbalanced class distributions. SegNet \cite{segnet}, another popular network, utilizes pixel-wise Softmax with Cross-Entropy loss for effective segmentation. DeepLab \cite{deeplab}, including its various iterations such as DeepLabv1, v2, v3, and v3+, employs pixel-wise Cross-Entropy loss and sometimes integrates Focal loss to enhance robustness against difficult examples. While, PSPNet (Pyramid Scene Parsing Network) \cite{pspnet} is designed to capture global context and also commonly used pixel-wise Cross-Entropy loss together with Focal loss. In the same context Efficient Neural Network (ENet) \cite{enet} is designed for real-time applications in semantic segmentation and often utilizes pixel-wise Cross-Entropy loss or Focal loss. BiFusion \cite{bifusion} focused on improving segmentation accuracy through a fusion of modalities and utilizes functions such as pixel-wise Cross-Entropy loss and structured losses, including Dice loss. HRNet (High-Resolution Network) \cite{hrnet} maintains high-resolution representations and typically employs pixel-wise Cross-Entropy loss or Focal loss to effectively segment objects. TriNet \cite{trinet} introduced a tri-level architecture to refine segmentation accuracy, generally deploying Cross-Entropy loss or Dice loss.

For ViT-based semantic segmentation many methods have been proposed using different loss functions. Each of these methods leverages specific loss functions according to their architectural design and the unique challenges presented by the datasets they aim to segment. For example, Attention U-Net \cite{attentionunet} enhances traditional U-Net structures with attention mechanisms and commonly utilizes Binary Cross-Entropy loss or Dice loss to improve focus on critical regions. 
SegFormer \cite{segformer} is another transformer-based model that has become popular for segmentation tasks, usually employing pixel-wise Cross-Entropy loss or Focal loss to tackle various segmentation challenges effectively. While, Segmenter \cite{segmenter} typically used Cross-Entropy loss for semantic segmentation tasks, which measures the difference between the predicted class probabilities and the actual classes for each pixel in the image. Swin-Transformer \cite{swinTR} is another method for semantic segmentation tasks, the Swin-Transformer often uses Cross-Entropy loss or can also incorporate Dice loss or Jaccard loss to better handle class imbalances, especially in cases where certain classes are underrepresented. DPT (Dense Prediction Transformer) \cite{dpt} typically leverages a combination of Cross-Entropy loss for segmentation along with specialized losses depending on the specific task (like L1 loss for depth estimation), focusing on both pixel-wise classification and dense prediction tasks.
MaskFormer \cite{maskformer} uses a Mask Classification loss which is effectively a variant of Binary Cross-Entropy loss, targeting the mask classification problem. It may also utilize Focal loss in some implementations to address class imbalance issues.
     
For instance segmentation, which involves identifying and segmenting individual objects, a combination of losses such as a pixel-wise Cross-Entropy loss for segmentation masks and Bounding Box Regression loss for object localization is often used as discussed in Object detection. These combined losses help in accurately identifying and separating each instance within an image. For example, Mask R-CNN extends object detection frameworks and combines several losses, including classification loss (Cross-Entropy), bounding box regression loss (Smooth L1 loss), and mask loss (pixel-wise Binary Cross-Entropy). 
  
Panoptic segmentation, which integrates both semantic and instance segmentation, typically employs a mix of the aforementioned loss functions. Additionally, the Panoptic Quality (PQ) metric is often used during evaluation. The loss functions in these tasks are designed to improve the model's precision in delineating and categorizing multiple objects and regions within the mage, ensuring high performance across the varied aspects of segmentation. For example, the Panoptic Feature Pyramid Network (Panoptic FPN) combines semantic and instance segmentation tasks, utilizing a mix of losses, including Cross-Entropy for semantic outputs and Focal loss or Dice loss as needed. 

\begin{table*}[t!]
    \centering
    \footnotesize
    \caption{Description of each loss function used in different discriminative tasks.}
    \label{descipriondescriminationloss}
    \begin{tabular}{|l|p{11cm}|}
        \hline
        \textbf{Loss Function} & \textbf{Description} \\
        \hline
\multicolumn{2}{|c|}{\textbf{Image Classfication}}
\\ \hline
        \hline
        Euclidean Loss & Measures the squared distance between predicted and true output vectors, commonly used in regression contexts. \\
        \hline
        Cross-entropy loss & Assesses the difference between predicted probabilities and actual class labels, widely used for classification tasks. \\
        \hline
        Attention-Target Loss & Enhances classification by focusing on relevant regions in the input, guiding the network to emphasize important features. \\
        \hline
        Ranking Loss & Evaluates the relative order of classes; useful in tasks like metric learning, where retaining rank information is crucial. \\
        \hline
        Kullback-Leibler Loss & Measures the divergence between two probability distributions, often used in scenarios involving knowledge distillation or when comparing predictions to true distributions. \\
        \hline \hline
\multicolumn{2}{|c|}{\textbf{Object detection}}
\\ \hline
        \hline

    Cross-Entropy Loss & Measures the difference between predicted probabilities and actual labels, used for classification tasks. \\
    \hline
    Mean Squared Error (MSE) & Average squared differences between predicted and actual values are commonly used in regression, including bounding box regression. \\
    \hline
    Smooth L1 Loss & Combines L1 and L2 characteristics; less sensitive to outliers than MSE, stabilizing bounding box regression. \\
    \hline
    Intersection over Union (IoU) Loss & Computes the overlap ratio between predicted and ground truth bounding boxes, maximizing this overlap. \\
    \hline
    Focal Loss & A variant of Cross-Entropy that emphasizes harder-to-classify examples, addressing the class imbalance in detection tasks. \\
    \hline
    Gaussian Wasserstein Distance & Captures the difference between predicted and ground truth bounding box distributions using Wasserstein distance. \\
    \hline
    Efficient IoU Loss & Optimizes traditional IoU Loss for better efficiency and stability while promoting accurate bounding box predictions. \\
    \hline\hline

\multicolumn{2}{|c|}{\textbf{Action recognition}}
\\ \hline \hline
        Cross-entropy loss & Measures the difference between predicted class probabilities and actual labels, commonly used for action classification tasks. \\
        \hline
        Contrastive Loss & Encourages similar actions to be close in the feature space while pushing dissimilar actions apart, used in metric learning and representation tasks. \\
        \hline \hline
\multicolumn{2}{|c|}{\textbf{Semantic Segmentation}}
\\ \hline \hline
        Binary Cross-Entropy & Measures the loss between predicted probabilities and actual binary labels for pixel classification. \\
        \hline
        Dice Loss & Focuses on maximizing the overlap between predicted and true segments, effective for imbalanced classes. \\
        \hline
        Pixel-wise Softmax with Cross-Entropy & Combines softmax for normalization with cross-entropy, used for multi-class segmentation. \\
        \hline
        Focal Loss & A variant of cross-entropy that emphasizes hard-to-classify examples and reduces the loss for well-classified ones. \\
        \hline
        Cross-entropy loss & Quantifies the difference between predicted class probabilities and actual labels, suitable for binary and multi-class tasks. \\
        \hline
        Pixel-wise Cross-Entropy & Computes cross-entropy for each pixel independently; ideal for multi-class classification at the pixel level. \\
        \hline
        Jaccard Loss & Measures the IoU between predicted and ground truth regions; useful in imbalanced scenarios. \\
        \hline
        L1 Loss & Calculates the mean absolute difference between predicted and actual pixel values, applied in regression contexts. \\
        \hline \hline
    \end{tabular}
    \label{tab:loss_functions}
\end{table*}

\textbf{Motion Detection:}
Motion detection focuses on identifying regions within video frames where movement occurs. Common loss functions for this task include Binary Cross-Entropy Loss and Foreground-Background Segmentation Loss. While, Binary Cross-Entropy Loss is used when the task involves classifying each pixel or region as either part of the moving object or the background. This loss function helps the model accurately distinguish moving regions from static ones. Foreground-background segmentation losses (such as Intersection over Union (IoU) or Dice Loss) ensures precise segmentation of moving objects by comparing predicted segmentation masks with ground truth masks, promoting better localization and delineation of motion areas.

Recently, several effective motion detection methods leveraging deep learning and background subtraction (BS) techniques have been proposed. For example, Kim et al. \cite{kim2019deep} demonstrated a deep learning-based background subtraction method using a convolutional neural network (CNN) to differentiate foreground from background, employing binary cross-entropy loss to classify pixels. Additionally, the authors in \cite{goyal2020deep} utilized a deep autoencoder for background representation, implementing reconstruction loss to differentiate moving objects from the background. CNNs have also been effectively applied in motion detection employing a CNN-based architecture to learn spatial features from video streams, optimizing their model using cross-entropy loss \cite{li2018motion}. The use of Long Short-Term Memory (LSTM) networks has also gained traction, where LSTMs model temporal dependencies in video sequences, employing sequence loss for enhancement of prediction accuracy \cite{jha2018deep}. More recently, Chen et al. introduced spatiotemporal convolutional networks that combine background modeling with deep learning techniques, optimizing the system using a multi-task loss that includes motion loss\cite{chen2023spatiotemporal}. Additionally, the authors in \cite{wang2021motion} combined background subtraction with optical flow techniques to enhance detection accuracy, employing a combination of Mean Squared Error and motion consistency loss. Motion consistency loss is used also for video generation methods.

\textbf{Object Tracking}
Object tracking aims to follow the trajectory of moving objects across consecutive video frames. The primary loss functions for tracking include Bounding Box Regression Loss (Smooth L1, logistic regression) \cite{BBR}, Re-Identification Loss (Contrastive loss, Triplet loss \cite{STTA}), and Temporal Consistency loss. Bounding Box Regression Loss (such as Smooth L1 loss or IoU loss) is used to refine the predicted bounding box coordinates, ensuring that the trackers accurately localize the object in each frame. Re-identification loss (ReID loss) helps the model maintain the identity of tracked objects across frames. This is often achieved by embedding learning techniques where the loss function minimizes the distance between embeddings of the same object in consecutive frames and maximizes the distance between different objects. Temporal Consistency Loss is used to ensure the smoothness and consistency of object trajectories over time. This loss can be formulated by penalizing abrupt changes in position or size of the tracked bounding boxes across consecutive frames, promoting more stable and realistic tracking.

\begin{figure*}[t!]
\centering
\includegraphics[width=.81\linewidth]{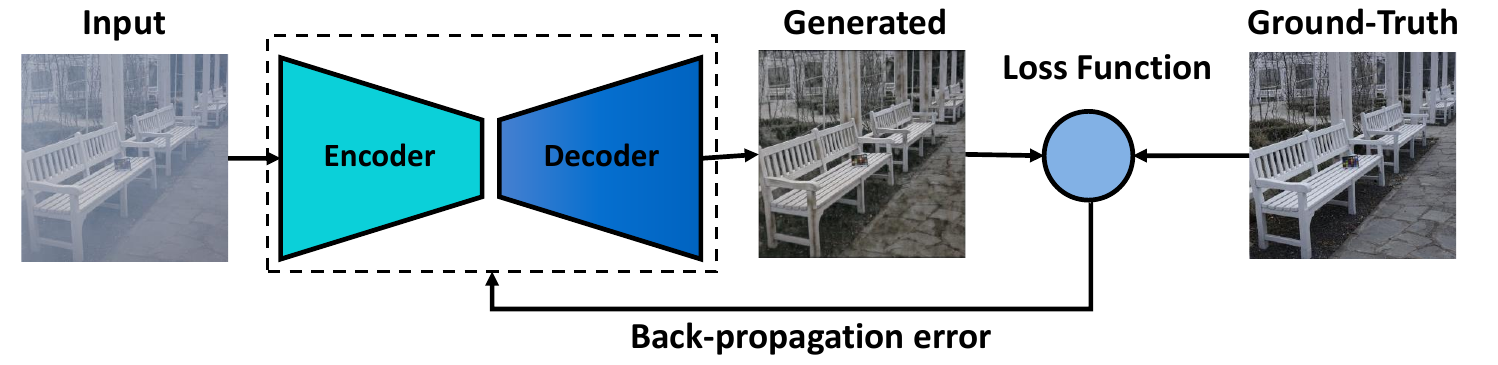}
\caption{Loss function in CNN-based models.}
\label{losscnn}
\end{figure*}

These loss functions work together to enhance the model's ability to accurately detect and consistently track moving objects over time, ensuring high performance in complex, real-world video data. They help the models to not only locate moving objects precisely but also maintain their identities and trajectories smoothly across frames. This is shown in the proposed methods using different deep learning techniques. One notable method is the Siamese Network approach, named SiamFC (Siamese Fully Convolutional), which uses similarity learning for online object tracking, employing cross-correlation loss to measure similarity between the template and the search area \cite{bertinetto2016fully}. Another influential method is the Deep SORT (Simple Online and real-time tracking), which combines deep appearance features with Kalman filtering for robust tracking, using Mahalanobis distance as part of its loss measurement \cite{wojke2017simple}. The Track R-CNN, presented by Voigtlaender et al., extends Mask R-CNN for tracking purposes, implementing a multi-task loss that includes Softmax Cross Entropy loss for classification, and Smooth L1 loss for bounding box regression \cite{voigtlaender2019track}. ATOM (Accurate Tracking by Overlap Maximization) focused on accurately estimating the overlap between the predicted and ground-truth bounding boxes, employing IoU (Intersection over Union) loss for refining tracking accuracy \cite{danelljan2019atom}. DiMP (Discriminative Model Prediction), optimized both localization and classification branches of tracking in a unified framework, using logistic regression loss to adapt the tracker dynamically \cite{bhat2019learning}. 

More recently, the OSTrack method introduced by Xu et al.  \cite{xu2021ostra} employs an online, end-to-end learning process that utilizes a multi-task loss to enhance tracking performance in dynamic environments. Additionally, the TransTrack method incorporated transformers into the tracking framework for better temporal feature extraction, using a combination of Focal, IoU, and L1 loss functions \cite{li2022transT}. The FairMOT framework efficiently addresses multi-object tracking by simultaneously detecting and tracking objects with a unified architecture, employing Re-Identification loss \cite{wang2020fairmot}.
In \cite{Diffusiontrack}, the authors proposed a diffusion model to track multiple objects using a set of loss functions including Focal, L1, and GIoU. While in \cite{Motiontrack} used L1 as a loss function in training the proposed model that consists of detection and re-identification techniques.

The loss functions used in each one from discriminative tasks are presented in Table \ref{descriminativeloss}, describing the technique used including statistical-based, CNN-based, and ViT-based techniques. A description of each loss function used is presented in Table \ref{descipriondescriminationloss}.

\subsection{Generative tasks}

Generative tasks in computer vision refer to processes aimed at creating new visual content or transforming existing images. These tasks are centered on models that learn the underlying distribution of data and utilize this understanding to generate original images, modify forms, or produce variations based on learned characteristics \cite{gen}. The field of generative tasks has gained significant traction, especially using deep learning architectures such as Convolutional Neural Networks (CNNs), Generative Adversarial Networks (GANs), Variational Autoencoders (VAEs), Vision transformers (ViT), and Diffusion models.

The primary goal of generative tasks is to enable machines to mimic the complexities of visual data, creating outputs that are indistinguishable from real images or producing variations that can enhance data diversity.

\begin{figure*}[t!]
\centering
\includegraphics[width=.81\linewidth]{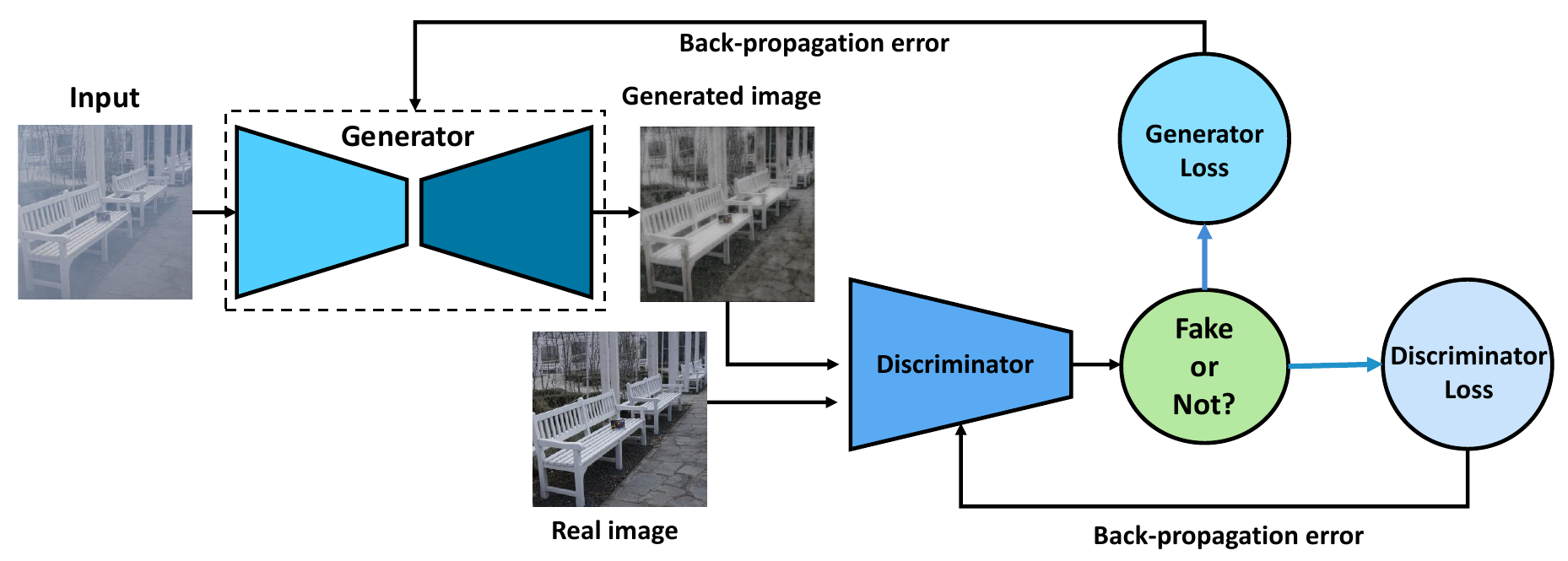}
\caption{Loss function in GAN-based models.}
\label{lossgan}
\end{figure*}

Loss functions are integral to the training of generative models, serving as the guiding metrics that quantify how well the generated outputs align with the desired targets as presented in Figures \ref{losscnn} and \ref{lossgan}. In generative tasks such as image-to-image, text-to-image, and audio-to-image generation, the choice of loss function significantly influences the quality and realism of the produced content. For instance, in image-to-image generation, pixel-wise and perceptual losses help ensure visual consistency and semantic relevance, while adversarial losses encourage the creation of realistic images through competition with discriminator networks. In the context of text-to-image generation, text-image similarity loss measures the alignment between generated visuals and input descriptions, enhancing accuracy in representation. Similarly, audio-to-image generation employs specialized loss functions that ensure the generated images meaningfully reflect the audio input. By optimizing these loss metrics, generative models can capture complex data distributions, leading to innovative applications in art, design, and multimedia content creation. As research progresses, the development and refinement of loss functions will play a critical role in advancing the capabilities of generative tasks across various domains. In this section a set of proposed methods for each one of these image generation tasks are discussed beside the used loss functions.  

\textbf{Text-to-Image Generation:}
In text-to-image generation, where the purpose is to generate images based on textual descriptions, Adversarial loss, and Text-Image Matching loss are used. While, adversarial loss, typically utilized in Generative Adversarial Networks (GANs), involves a discriminator trying to differentiate between real and generated images and a generator trying to produce images indistinguishable from real ones. This loss helps improve the realism of generated images. Text-image matching loss is used to ensure that the generated image accurately reflects the given text description, often implemented via pair-wise ranking losses or similarity losses that align the text and image embeddings.

Recent advancements in text-to-image generation have led to several innovative techniques. Generative Adversarial Networks (GANs) have been extensively utilized, with the improved model known as AttnGAN proposed by Xu et al. \cite{xu2018attngan}, which incorporates attention mechanisms to enhance the generated images based on descriptive text inputs, utilizing adversarial loss alongside L1 loss. Another text-to-image method is DALL-E \cite{rameswar2021dall} which employs a transformer-based architecture capable of generating high-quality images from textual descriptions, typically using categorical cross-entropy loss for training. Moreover, the CLIP (Contrastive Language-Image Pretraining) framework introduced demonstrates the effectiveness of aligning text and image representations, using contrastive loss to enhance the image generation process \cite{radford2021learning}. More recently, the Stable Diffusion model emerged, facilitating high-quality images based on complex text prompts through diffusion processes, by using noise prediction loss during training \cite{rombach2022high}. The text-to-image generation approach via Diffusion Models was further refined by Dhariwal and Nichol \cite{dhariwal2021diffusion}, which generates high-fidelity images by gradually denoising samples and employs noise prediction loss. 

Additionally, the method by Song et al. \cite{song2021generative}, known as Guided Diffusion, enhances the generation process by conditioning various input types, utilizing a guided approach for diffusion loss. Furthermore, the method presented in \cite{park2021stylegan2}, called StyleGAN2-ADA, extends the original StyleGAN framework with adaptive data augmentation, improving the quality of image generation with novel perceptual loss functions. Finally, the method called GLIDE combined diffusion models with a natural language understanding component to generate images from textual cues, incorporating variational autoencoder loss \cite{nichol2022glide}.

\begin{table*}[t!]
\centering
\footnotesize
\caption{Loss functions used in image generation techniques}
\label{genlossmethods}
\begin{tabular}{|p{3cm}|p{4.3cm}|p{1.5cm}|p{6cm}|}
\hline
\textbf{Task} & \textbf{Method} & \textbf{Technique} &\textbf{Loss Function} \\
\hline
\hline
\multirow{5}{*}{\textbf{Text-to-Image}} & AttnGAN \cite{xu2018attngan}  & GAN &Adversarial, L1 \\\cline{2-4}
\multirow{5}{*}{\textbf{Generation}}& DALL-E \cite{rameswar2021dall} & ViT & Categorical Cross-Entropy \\\cline{2-4}
& CLIP \cite{radford2021learning} & ViT &Contrastive \\\cline{2-4}
& Stable Diffusion \cite{rombach2022high}& Diffusion & Noise Prediction \\\cline{2-4}
& Dhariwal et al. \cite{dhariwal2021diffusion}& Diffusion & Noise Prediction \\\cline{2-4}
& Guided-Diffusion \cite{song2021generative}&Diffusion & Diffusion \\ \cline{2-4}
&StyleGAN2-ADA \cite{park2021stylegan2}&GAN&Perceptual  \\ \cline{2-4}
&GLIDE  \cite{nichol2022glide}&Diffusion &Variational autoencoder \\ \hline\hline

\multirow{7}{*}{\textbf{Image-to-Image}} & Pix2Pix \cite{isola2017image} & GAN & L1, Conditional  Adversarial \\\cline{2-4}
\multirow{7}{*}{\textbf{Generation}}& CycleGAN \cite{zhu2017unpaired}&GAN & Cycle Consistency \\\cline{2-4}
& SPADE \cite{park2019semantic}& CNN  & Adversarial, Perceptual\\\cline{2-4}
& StarGAN \cite{choi2020stargan}& GAN & Cycle Consistency, Pixel-wise\\\cline{2-4}
& StyleGAN/StyleGAN2 \cite{karras2019style,karras2021analyzing}& GAN & Perceptual \\\cline{2-4}
&Saito et al. \cite{saito2019image} & GAN &GAN , Perceptual \\\cline{2-4}
& GauGAN \cite{park2019gaugan} & GAN &Adversarial, Context \\\cline{2-4}
&Huang et al. \cite{itr2024} & GAN &Cycle consistency, Adversarial \\\cline{2-4}
&CLUIT \cite{CLUIT} & GAN &Adversarial, Contrastive \\\cline{2-4}
&Diffi2i \cite{Diffi2i} & Diffusion &Perceptual, Adversarial, L1\\\hline\hline
\multirow{5}{*}{\textbf{Audio-to-Image}} & SoundNet \cite{aytar2016soundnet}& CNN& MSE, Adversarial \\\cline{2-4}
\multirow{5}{*}{\textbf{Generation}}& Tzinis et al. \cite{tzinis2020audio} & GAN&Adversarial, Perceptual \\\cline{2-4}
& Chen et al. \cite{chen2020audio} &CNN &Reconstruction, KL Divergence \\\cline{2-4}
& Sound2Visual \cite{baird2021sound2visual} &GAN &Adversarial, Pixel-wise \\\cline{2-4}
& Zhao et al. \cite{zhao2022audiovisual}&Diffusion   & Noise Prediction \\\cline{2-4}
&CMCRL \cite{CMCRL}& GAN& Adversarial \\\cline{2-4}
&MACS \cite{MACS}& CNN, ViT& Ranking, Contrastive\\ \hline\hline

\multirow{5}{*}{\textbf{Image Inpainting}} &Partial Convolution  \cite{liu2018image}&CNN&L1, Style , Perceptual \\\cline{2-4}
 &CTN \cite{21-1}& CNN, ViT& Reconstruction, Perceptual, Adversarial, Style \\\cline{2-4}
&MAT \cite{21-4}& CNN, ViT&Perceptual, Adversarial\\\cline{2-4}
&  T-Former \cite{22-2} & ViT &   Reconstruction, Perceptual, Adversarial, Style\\\cline{2-4}
& TransCNN-HAE   \cite{22-7}& CNN, ViT &  Reconstruction, Perceptual, Style  \\\cline{2-4}
&  CoordFill \cite{23-11}  & CNN, ViT &  Perceptual, Adversarial, and Feature Matching  \\\cline{2-4}
&  CMT \cite{23-16}  & CNN, ViT &  Reconstruction, Perceptual, Adversarial  \\\hline\hline

\multirow{5}{*}{\textbf{Image Dehazing}} & Araji et al. \cite{da1} &CNN& L1\\\cline{2-4}
                &FFA-Net\cite{d1} & CNN& L1\\\cline{2-4}
               & DehazeFormer\cite{d2} & ViT& L1\\\cline{2-4}
               & Li et al. \cite{d3} & CNN-ViT& SSIM, MSE\\\cline{2-4}
               & CycleGAN \cite{d4} & GAN & Cycle Consistency , SSIM, Pixel-wise mean \\\cline{2-4}
                & HSMD-Net \cite{d5} & CNN & SmoothL1 , Perceptual, Adversaria, MS-SSIM \\\hline
\end{tabular}
\end{table*}

\textbf{Image-to-Image Generation:}
For image-to-image translation tasks, such as converting sketches to images, day-to-night transformations, or also transforming images we can find a set of loss functions including Adversarial loss, Cycle-Consistency loss, GAN loss, Perceptual loss, and Pixel-wise Losses (like L1 or L2 loss) that commonly used. Adversarial Loss promotes the generation of realistic images, while Cycle-Consistency Loss is critical in models like CycleGANs, ensuring that an image translated to another domain and then back again retains its original content. Pixel-wise Losses help in maintaining finer details by comparing corresponding pixel values between the generated and target images.

Recent advancements in image-to-image generation (or translation) have produced several innovative techniques. From these methods, we can find  Pix2Pix that utilized conditional Generative Adversarial Networks (cGANs) to learn a mapping between paired images, employing L1 loss to preserve the image details along with adversarial loss for realistic output generation \cite{isola2017image}. Following this, CycleGAN is another famous method that provides a framework for unpaired image-to-image translation, incorporating cycle consistency loss to ensure that translations remain reversible \cite{zhu2017unpaired}. Another significant approach, named SPADE (Spatially-Adaptive Normalization) enhanced semantic image synthesis by applying spatially adaptive normalization layers, relying on adversarial loss and perceptual loss \cite{park2019semantic}. 

More recently, the method known as StarGAN expands the scope of image-to-image translation to handle multi-domain image synthesis while using cycle consistency loss and pixel-wise loss \cite{choi2020stargan}. StyleGAN and its extension StyleGAN2 developed by Karras et al., excel in generating high-quality, high-resolution images and can also be adapted for image translations effectively, using perceptual loss functions that enhance the quality of generated images \cite{karras2019style, karras2021analyzing}. Moreover, the proposed method by Saito et al. \cite{saito2019image}, focuses on learning latent representations to facilitate intuitive image translations, employing a combination of GANs and perceptual loss. While GauGAN model is a method that allows users to create photorealistic images from segmentation maps using a combination of adversarial loss and context loss to maintain the spatial context in the translations \cite{park2019gaugan}. In \cite{itr2024}, the authors used cycle consistency loss and two adversarial loss functions to generate underwater images based on the CycleGan model. Based also on the GAN model, the authors in \cite{CLUIT} proposed an image-to-image translation named CLUIT  using adversarial and contrastive loss functions. Wile in the proposed diffusion-based method proposed in \cite{Diffi2i} incorporates perceptual, adversarial, and L1 loss functions to train their model.

\textbf{Audio-to-Image Generation}
For audio-to-image generation, the goal is to generate visuals based on audio inputs (such as generating images from speech or music). The loss functions used for that includes Adversarial loss, Reconstruction loss, and Consistency loss. Adversarial Loss ensures the generated images are realistic. Reconstruction Loss compares the generated image to a target image or measures how well the generated image can be used to reconstruct the original audio. Audio-Visual Consistency Loss ensures the generated image is contextually aligned with the audio input, often involving aligning embeddings from audio and image modalities.

These loss functions collectively guide the models to generate high-quality, contextually relevant images, ensuring that the outputs are not only visually appealing but also appropriately aligned with their respective inputs, whether textual, visual, or auditory.



Recent advancements in audio-to-image generation have led to notable techniques that link between auditory and visual content. For example, the method proposed in \cite{aytar2016soundnet}, known as SoundNet, learns to generate visual representations from audio inputs by employing a convolutional neural network (CNN) trained with both audio and corresponding visual data, primarily using a combination of mean squared error (MSE) loss and adversarial loss. While in \cite{tzinis2020audio}, Tzinis et al. introduced a model that generates sketches from audio signals, utilizing a conditional GAN framework and employing adversarial loss combined with a perceptual loss to enhance the quality of the generated sketches. Another method proposed by Chen et al. \cite{chen2020audio} demonstrated the effectiveness of jointly modeling audio and visual information by leveraging a variational autoencoder (VAE) architecture that utilizes reconstruction loss and Kullback-Leibler divergence as part of the training process. Furthermore, Baird et al. \cite{baird2021sound2visual} implemented a model named Sound2Visual to generate detailed visual images from sound input by adopting a GAN architecture alongside adversarial loss and pixel-wise loss to retain the fidelity of the generated images. While, the proposed approach in  \cite{zhao2022audiovisual} focused on converting complex auditory scenes into images through a two-stage process utilizing a diffusion model, employing noise prediction loss to refine image generation. While in \cite{CMCRL}, Adversarial loss is used to implement the CMCRL method which is a GAN-based method. Two other loss functions are incorporated to train the CNN-based method named MACS.

\begin{table*}[t!] 
    \centering
    \caption{Description of loss functions used in generative tasks }
    \label{genloss}
    \begin{tabular}{|l|p{10cm}|}
        \hline
        \textbf{Loss Function} & \textbf{Description} \\
        \hline

        MSE Loss & Measures the mean squared difference between predicted and actual image values, ensuring pixel-level accuracy. \\   \hline
        
        L1 Loss & Measures the mean absolute difference between predicted and actual pixel values, encouraging accuracy in image generation. \\  \hline
        
        Categorical Cross-Entropy Loss & Calculates the difference between predicted class distributions and true labels, typically used in classification contexts for image features. \\  \hline
        
        Style Loss & Evaluates the similarity in style between the generated image and a reference image, focusing on texture and color distribution. \\
        \hline
        Reconstruction Loss & Assess how well the generated image matches the input information, ensuring important details are preserved. \\
        \hline
    
        Adversarial Loss & Challenges the generator to produce realistic images by evaluating its outputs against those of a discriminator. \\
        \hline
        Perceptual Loss & Evaluate image quality based on high-level features from pre-trained networks, emphasizing perceptual similarity rather than pixel accuracy. \\
        \hline        
        
        Contrastive Loss & Ensures that image representations align closely with text embeddings, useful in joint embedding spaces. \\
        \hline

        Variational Autoencoder Loss & Combines reconstruction loss with a regularization term to ensure meaningful latent spaces, fostering diverse image outputs. \\
        \hline

        Conditional Adversarial Loss & Guides the generator by conditioning it on extra information (e.g., class labels) to produce images relevant to that context. \\
        \hline

        Cycle Consistency Loss & Aim to retain content and structure across transformations between domains, enforcing consistency in generated outputs. \\
        \hline
        Context Loss & Focuses on preserving the semantic context and spatial structure of images during generation to ensure contextual accuracy. \\ \hline

        KL Divergence Loss & Measures how one probability distribution diverges from a second, often used in scenarios involving distribution matching. \\
        \hline
        Noise Prediction Loss & Focuses on predicting and minimizing noise in the generated output, improving the quality of the final image. \\
        \hline

        Diffusion Noise Prediction Loss & Used in diffusion models to predict noise and refine images iteratively, enhancing generation quality. \\
        \hline
        Diffusion Loss & Related to diffusion processes, measure the difference between image outputs at different steps in progressive refinement. \\
        \hline
    \end{tabular}
    \label{tab:loss_functions}
\end{table*}  

\textbf{Image Inpainting:}

Image inpainting is a technique used in computer vision and image processing to restore or fill in missing or damaged parts of an image. The goal is to reconstruct the missing areas to get a realistic completed image. Different deep learning techniques are used including CNN-based, and ViT-based methods for image and video inpainting. Multiple loss functions are employed to guide realistic image generation \cite{inp1}. Typically, these methods use a combination of loss functions to accommodate the diverse objectives they aim to achieve. Commonly used loss functions in image inpainting include Mean Absolute Error Loss (L1), Adversarial Loss, Perceptual Loss, Reconstruction Loss, Style Loss, and Feature Map Loss.

These and other loss functions are implemented across methods utilizing CNN, ViT, or GAN techniques \cite{inp2}. For instance, the Partial Convolution method
\cite{liu2018image} uses masked convolutions to prioritize valid pixel areas during training, effectively addressing irregular hole inpainting. It frequently employs L1 loss for pixel precision, style loss for texture alignment, and perceptual loss for maintaining feature integrity. Similarly, the Contextual Transformer Network (CTN) \cite{21-1} uses a ViT-based framework to handle relationships between corrupted and intact regions, employing a combination of Reconstruction, Perceptual, Adversarial, and Style loss functions.

Additionally, the Mask-aware Transformer (MAT) \cite{21-4} is designed for efficiently inpainting large holes in high-resolution images, drawing on transformers and convolutions for effective long-range interaction and processing, respectively. Predominantly, Perceptual and Adversarial loss functions are utilized. The T-former method, another ViT-based approach, offers long-range modeling with lower computational demand, using Reconstruction, Perceptual, Adversarial, and Style losses. Meanwhile, TransCNN-HAE \cite{22-7} combines both ViT and CNN networks to address varied image damage using Reconstruction, Perceptual, and Style losses.
CoordFill \cite{23-11} is another method that employs an attentional fast Fourier convolution (FFC)-based generation network that captures extensive reception fields through lower-resolution encoding, enhancing high-frequency texture synthesis via continuous position encoding. For CoordFill, Perceptual, Adversarial, and Feature Matching losses are used. The Continuous Mask-aware Transformer (CMT) \cite{23-16} trains by using Reconstruction, Perceptual, and Adversarial loss functions to represent error amounts with continuous masks.

Incorporating a range of loss functions in image inpainting and translation tasks, like image generation and segmentation, can enhance both the visual and semantic quality of results. These loss functions can be categorized into contextual-based, style-based, and structure-based groups. Contextual-based losses maintain semantic information, making inpainted regions seamlessly merge with surrounding areas. L1 and Reconstruction losses fall into this category. Style-based losses capture high-level semantic qualities rather than focusing on detailed pixels, targeting textures and artistic styles via feature map statistics. This category includes perceptual, style, and adversarial losses. Lastly, structural-based losses focus on contextual harmony, emphasizing the retention of structural coherence and surrounding content integrity. 

\begin{figure*}[t!]
\centering
  \begin{tabular}[b]{c}
    \includegraphics[width=1\linewidth]{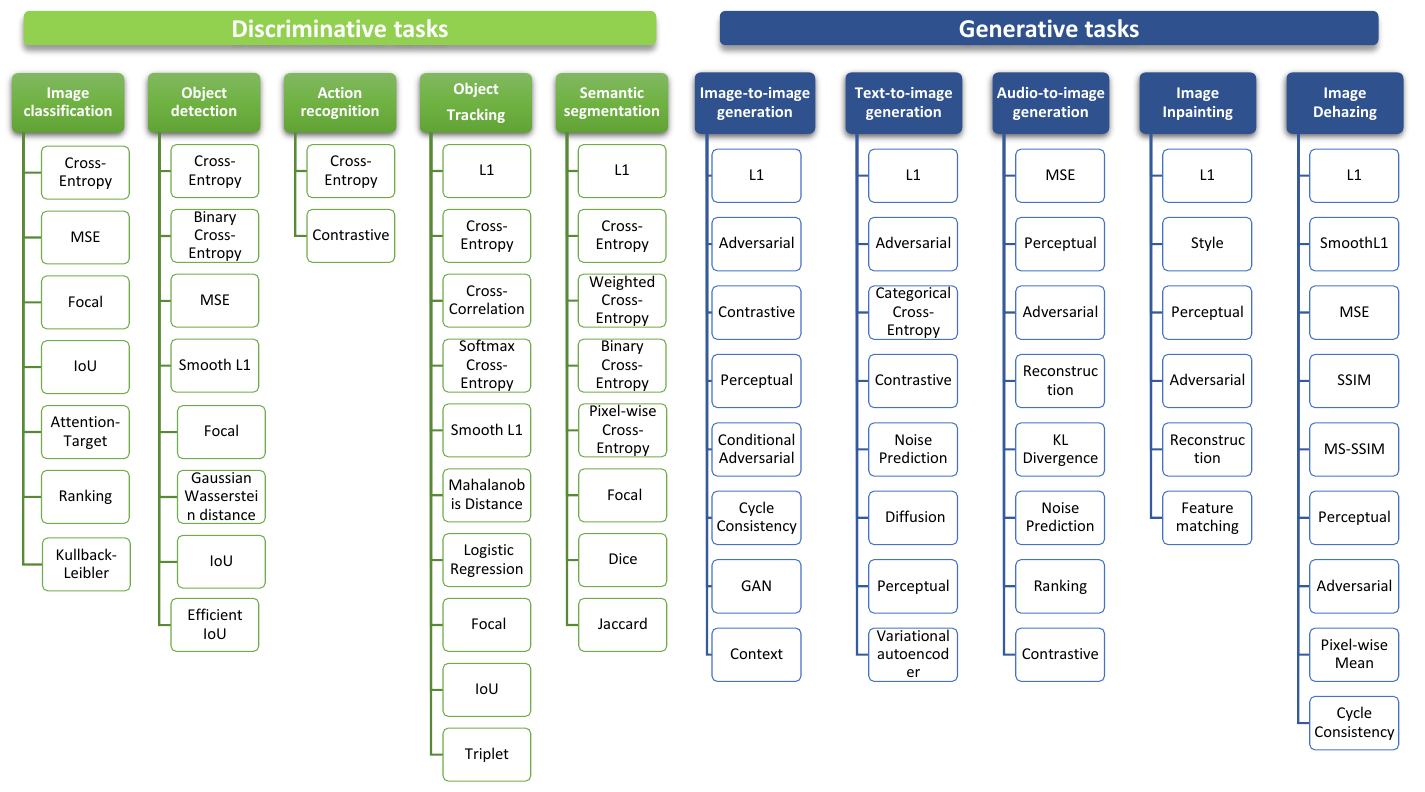}\\
  \end{tabular}
  \caption{{Loss function used in various computer vision (Discriminative, generative) tasks.}}
\label{allgentasksandlosses}
\end{figure*}


\textbf{Image dehazing:}
Image and video dehazing is a crucial area of research in computer vision that aims to enhance the visual quality of hazy images and videos by removing or reducing haze effects. Various types of image haze can affect photographs, including Atmospheric Haze, Foggy Haze, and Haze from Pollution. While, atmospheric Haze is caused by the scattering of light particles in the atmosphere, leading to decreased contrast and color saturation \cite{da1}. Foggy Haze refers to a dense atmospheric condition that creates a misty, diffused appearance in images. Haze from Pollution results from airborne pollutants, such as smog or industrial emissions, which can impart a hazy look to photographs, often giving them a yellowish or brownish tint and reducing clarity. Additionally, other types of haze can result from environmental factors like heat, rain, and varying levels of light \cite{da2}. Each haze type can influence the mood and atmosphere of an image, often requiring different techniques for effective minimization or enhancement of its effects.

In image dehazing, many important works have been done to improve the quality of hazy images by using different algorithms and datasets. For example, Araji et al. \cite{da1} proposed a model to conserve the high quality of the dehazed images using a multi-scale representation network. In this method, $L_1$ is employed as a loss function. In \cite{da2}, the author proposed HSMD-Net, which includes a hierarchical slice information interaction module (HSIIM) to enhance feature representation through intra-layer and inter-layer correlation, and a multi-layer cooperative decoding reconstruction module (MCDRM) to fully leverage feature information from all decoding layers, improving color and texture restoration. As loss function, the authors combine four loss functions including SmoothL1, Perceptual, Adversaria, and Multiscale Structural Similarity losses.

While the authors in \cite{d1} introduced an end-to-end feature fusion attention network (FFA-Net) for restoring haze-free images. It consists of a Feature Attention (FA) module that combines Channel and Pixel Attention to enhance representation. Also, a block structure with Local Residual Learning that prioritizes significant information. In addition an Attention-based Feature Fusion (FFA) structure that adaptively learns feature weights, emphasizing important features while retaining shallow information for deeper layers. While $L_1$ is used as a loss function to train FFA-Net. Using the same, loss function but with a ViT-based network instead of a CNN-based network, the authors in \cite{d2} proposed DehazeFormer a ViT-based model inspired by Swin-Transformer, which includes improvements like a modified normalization layer, a new activation function, and spatial information aggregation, while also using the $L_1$  loss function to enhance training outcomes. Combining CNN and ViT networks, the authors in \cite{d3} aim to be more effective in various hazy conditions by using a CNN-based and ViT-based network to capture local features and global features. While $SSIM$ and  $MSE$ loss functions are used to train the proposed model. 

Other than CNN and ViT, many methods have been proposed based on GAN which used different loss functions. The authors in \cite{d4}  have utilized advanced models like CycleGAN to tackle image dehazing, training their approach using cycle consistency, SSIM, and pixel-wise mean loss functions.

An overall idea about the used loss function in each task has been summarized in tables and a Figure.
Table \ref{genlossmethods} summarizes the loss functions used for all types of image generation. Table \ref{genloss} provides a brief description of each of these loss functions used in generating real images. In addition, Figure \ref{allgentasksandlosses} illustrates each task with all related loss functions. 




\subsection{Loss function per architecture}
The fundamental purpose of a loss function is the same in both Convolutional Neural Networks (CNNs) and Vision Transformer (ViT) models: to quantify the difference between the model's predictions and the actual target values. This crucial metric guides the optimization process during the training phase, allowing the models to learn from errors and improve over time. Furthermore, both CNNs and ViTs utilize backpropagation to compute the gradients of the loss concerning the model parameters. This capability allows for systematic adjustments to the weights and biases, making the learning process effective across different architectures. In terms of the types of loss functions used, both models can employ similar options depending on the task.
\begin{table*}[t!]
    \centering
    \caption{Comparison of Loss Functions in CNN-based and ViT-based methods }
    \label{compcnnvit}
    \begin{tabular}{lp{6cm}p{6cm}}
        \toprule
        \textbf{Aspect}                              & \textbf{CNN (Convolutional Neural Networks)}                   & \textbf{ViT (Vision Transformers)}                             \\ \midrule
        Input Processing                             & Processes whole images using convolutional filters to capture local features.                       & Segments images into patches, relying on self-attention for global dependencies.         \\ \hline
        Impact of Input Representation                & Generates predictions based on local patterns, impacting loss interpretation.                       & Understands global context, influencing loss interpretation differently.                    \\ \hline
        Dataset Size                                 & Generally effective with smaller datasets.                                                        & Often requires larger datasets for optimal performance.                                   \\  \hline
        Sensitivity to Hyperparameters                & Relatively less sensitive to hyperparameters and initialization strategies.                        & More sensitive to hyperparameters, affecting optimization and training dynamics.          \\ \hline
        Performance on Tasks                         & Excels in tasks like image classification and segmentation.                                       & Performs better in scenarios where capturing global context is essential.                \\ \hline
        Selection of Loss Functions                   & Utilization of various loss functions based on the task requirements.                                   & Chooses loss functions that are often optimized for global context understanding.       \\ \hline
        Overall Strategy of Loss Functions           & Consistent strategy of guiding optimization through error quantification.                         & Same strategic purpose; influenced by architecture-specific characteristics.               \\ \hline
        Application Differences                       & Requires less specialized approaches due to local feature focus.                                  & Needs tailored approaches for loss function application based on architecture and tasks.  \\
        \bottomrule
    \end{tabular}
\end{table*}

Besides these similarities, there are some differences due to the distinct architectures and processing methodologies inherent to CNNs and ViTs. For instance, CNNs typically process whole images as input, leveraging convolutional filters to capture local features and spatial hierarchies. In contrast, ViT models first segment images into patches, and their performance relies heavily on self-attention mechanisms to understand global dependencies among these patches. This inherent difference in input representation can influence not only the way predictions are generated but also the interpretation of the loss calculated during training.

Additionally, the training dynamics between the two types of models can vary significantly. ViTs often require larger datasets and may be more sensitive to hyperparameter settings and initialization strategies than CNNs. Such factors can impact the optimization process and how effectively different loss functions contribute to learning. Moreover, the performance of various loss functions may differ depending on the model architecture. For example, CNNs may excel in tasks like image classification and segmentation, while ViTs may have an advantage in scenarios where capturing global context is essential. This leads practitioners to select different loss functions or training strategies based on the expected performance characteristics of the model they are using.

The benefit of using loss functions remains consistent between CNNs and ViT-based models, the differences in architecture, input representation, and training dynamics mean that the application and effects of these loss functions can vary significantly. The choice of a specific loss function and its overall impact can be influenced by the unique characteristics of the model and the particular nature of the tasks being performed. a summarization of these differences is shown in Table \ref{compcnnvit}.

\section{Loss function for tabular data prediction}

\begin{figure*}[t!]
\centering
  \begin{tabular}[b]{c}
    \includegraphics[width=.8\linewidth]{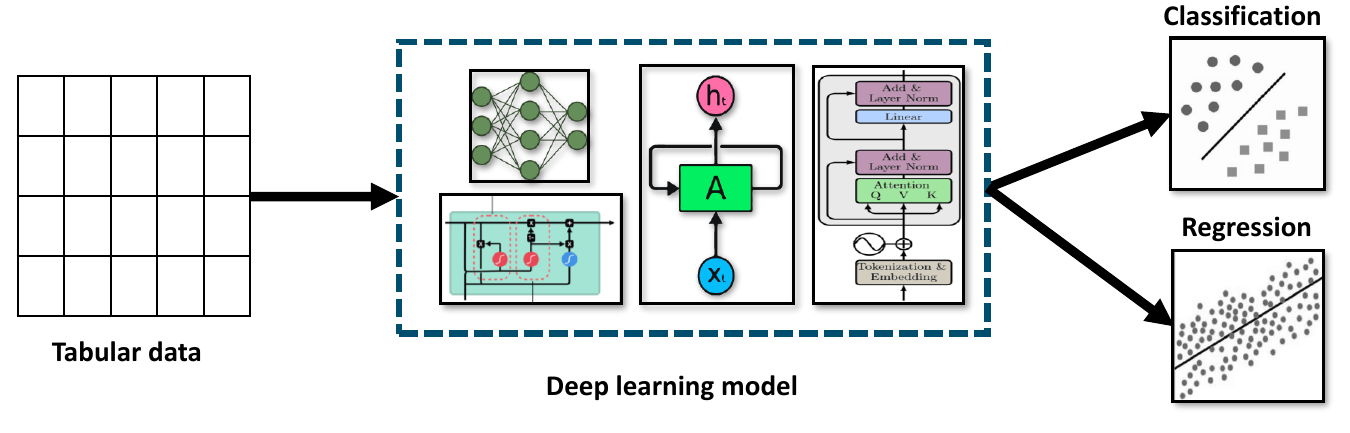}\\
  \end{tabular}
  \caption{{Tabular data prediction process.}}
\label{tab}
\end{figure*}

Tabular data prediction is a fundamental task in machine learning and data analysis, where the goal is to make predictions, classification or regression, based on structured data organized in rows and columns as presented in Figure \ref{tab}. In tabular datasets, each row typically represents a data point or an observation, while columns represent features or variables associated with that data \cite{tdr1}. This type of data is commonly found in databases, spreadsheets, and structured tables, making it a prevalent form of data in various domains such as finance, healthcare, marketing, and more.

The process of tabular data prediction involves training machine/deep learning models on historical data to learn patterns and relationships that can be used to make predictions on new, unseen data \cite{tdr2}. These predictions can range from simple tasks like predicting a numerical value in regression problems to more complex tasks like classifying data into different categories in classification tasks. Popular machine learning algorithms used for tabular data prediction include linear regression, decision trees, random forests, support vector machines, and neural networks \cite{tdr3}.

In tabular data predictions, selecting an appropriate loss function is crucial for effectively training machine learning models. Commonly used loss functions for tabular data prediction tasks include Mean Squared Error (MSE) and Mean Absolute Error (MAE) for regression problems, where the goal is to predict continuous values. These loss functions quantify the deviation between predicted values and actual targets, measuring model performance. For classification tasks on tabular data, Cross-Entropy-based loss functions such as Binary Cross-Entropy or Categorical Cross-Entropy are often employed to evaluate the discrepancy between predicted class probabilities and true labels. Choosing the right loss function for tabular data predictions is essential for guiding the learning process and optimizing model parameters to achieve accurate and reliable predictions on structured datasets.

In this context, many methods have been proposed using different techniques \cite{tdr1} from statistical to machine/deep learning. For example, \textbf{Linear regression} is a foundational approach that models the relationship between a dependent variable and one or more independent variables, commonly utilizing Mean Squared Error (MSE) as its loss function. \textbf{Decision trees} provide a non-parametric method for both classification and regression tasks, using MSE for regression tasks and Gini impurity for classifications.

Also, we can find \textbf{Random Forest} which utilizes an ensemble of decision trees to improve prediction accuracy and robustness, employing the average MSE for regression problems or Cross-Entropy Loss for classification tasks. \textbf{Gradient Boosting Machines (GBM)} represent another powerful ensemble technique, sequentially building models to minimize a differentiable loss function, typically opting for MSE in regression scenarios or Log Loss for classifications.

\textbf{XGBoost} is a popular variant of GBM, designed for increased speed and performance on larger datasets, and it can utilize logarithmic loss or MSE depending on the prediction task. On the deep learning side, Deep Neural Networks (DNN) can capture complex relationships in tabular data, using MSE for regression or Binary Cross-Entropy loss for classification.

Two notable gradient-boosting frameworks emerging in recent years are LightGBM and CatBoost. LightGBM focuses on speed and efficiency, integrating gradient-based one-side sampling and histogram-based algorithms. While CatBoost is specifically optimized to handle categorical data directly, employing Log Loss or MSE where applicable. Each of these methods offers distinct advantages suited to various types of tabular data, showcasing the diversity of approaches available in this domain.

Recent advancements focused on deep learning techniques can effectively capture complex relationships within the data. For these methods we can find Deep Neural Networks (DNN), which consist of multiple layers of neurons that can model nonlinear relationships, typically using Mean Squared Error (MSE) for regression tasks or Binary Cross-Entropy for classification tasks \cite{goodfellow2016deep}.
Another significant approach is Gradient Boosting with Neural Networks (XGBoost) that effectively integrates deep learning principles with boosting techniques for enhanced predictive performance on structured data, employing log loss or MSE depending on the prediction context \cite{chen2016xgboost}. TabNet is another recent architecture that leverages attention mechanisms to process tabular data more efficiently, optimizing with a combination of cross-entropy loss and mean squared error as needed for different tasks \cite{arik2019tabnet}.
Entity Embeddings of Categorical Variables introduced in \cite{din2020deep} is another innovative application of deep learning to tabular data. It uses neural networks to convert categorical variables into dense vectors, effectively capturing relationships between categories, often optimizing with cross-entropy loss for classification tasks \cite{din2020deep}. Additionally, AutoGluon \cite{erez2021autogluon} provides a high-level interface that applies ensemble deep learning techniques automatically to tabular data, showcasing robust performance across various tasks while employing multi-task loss functions to optimize different components of the model \cite{erez2021autogluon}.
\begin{table*}[t!]
\centering
\caption{Summary of tabular data prediction methods and the used loss functions.}
\label{predforcast}
\begin{tabular}{|p{2cm}|p{2.5cm}|p{2cm}|p{2.5cm}|p{6cm}|}
\hline
\textbf{Type of Data} &  \textbf{Task} & \textbf{Class/Reg} & \textbf{Technique} &\textbf{Loss Function} \\ \hline

& DNN \cite{goodfellow2016deep}& Regression &DNN& MSE \\ \cline{2-5}
& DNN \cite{goodfellow2016deep}& Classification &DNN&  Binary Cross-Entropy\\ \cline{2-5}
&XGBoost \cite{chen2016xgboost} & Classification&ANN &Log or MSE \\ \cline{2-5}

\multirow{1}{*}{\textbf{Tabular Data}}&TabNet \cite{arik2019tabnet}& Class, Reg & Transformer & Softmax Cross-Entropy (Class),  MSE (Reg) \\ \cline{2-5}
\multirow{1}{*}{\textbf{Prediction}}&Din et al. \cite{din2020deep}& Classification&ANN & Cross-Entropy (class) \\ \cline{2-5}
&AutoGluon-Tabular\cite{erez2021autogluon}& Class, Reg&ANN& Log, MSE \\ \cline{2-5}
&CatBoost \cite{prokhorenkova2018catboost}& Class, Reg&Gradient Boosting&  Log , Zero-one \\ \cline{2-5}
&LightGBM \cite{ke2017lightgbm}& Class &Decision Tree & Various loss functions \\ \hline
\end{tabular}
\end{table*}
More recently, CatBoost, which incorporates gradient boosting techniques optimized specifically for categorical features, uses various loss functions like log loss or MSE based on the nature of the prediction \cite{prokhorenkova2018catboost}. While, LightGBM leverages a histogram-based algorithm designed for efficiency and speed on large datasets, benefiting from deep learning techniques to optimize the loss functions used in gradient boosting \cite{ke2017lightgbm}. 

A summarization of the selected methods with the loss function used is provided in Table \ref{predforcast}.

\section{Loss function for time series forecasting}

\begin{figure*}[t!]
\centering
  \begin{tabular}[b]{c}
    \includegraphics[width=.8\linewidth]{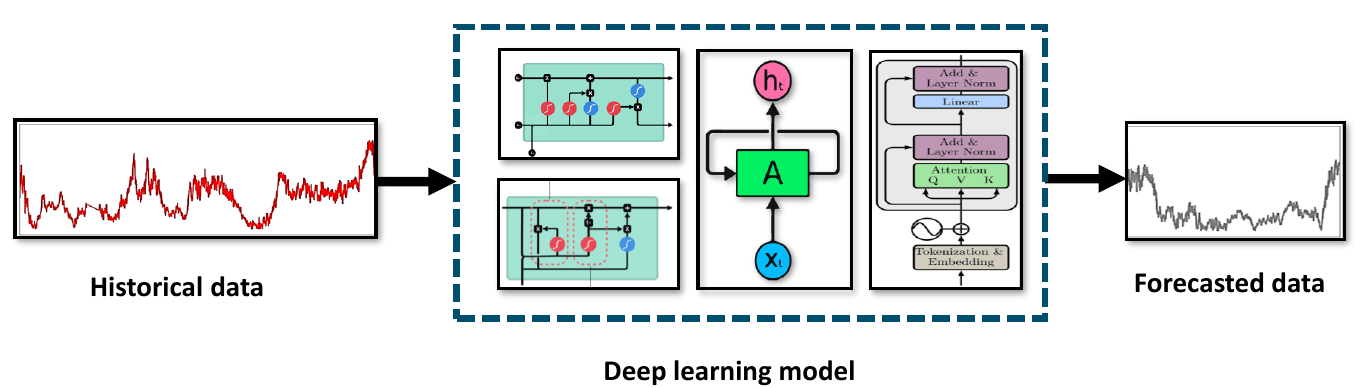}\\
  \end{tabular}
  \caption{{Time series forecasting learning process.}}
\label{times}
\end{figure*}

Predicting future trends, events, or behaviors based on historical data is a common yet crucial task in many domains. Time series forecasting, in particular, involves forecasting future values of a sequence of data points based on patterns and dependencies observed in past data. As presented in Figure \ref{times}, deep learning models, such as recurrent neural networks (RNNs) and their variations like Long Short-Term Memory (LSTM) and Gated Recurrent Unit (GRU), offer the capability to capture long-term dependencies in sequential data \cite{ts1}. These models excel at capturing intricate patterns in time series data, making them suitable for tasks like stock price forecasting, weather forecasting, energy demand forecasting, and more.

One of the key advantages of using deep learning for time series forecasting is its ability to automatically learn and extract features from raw temporal data without the need for manual feature engineering. This can be particularly beneficial when dealing with large-scale, high-dimensional time series datasets where identifying relevant features manually might be challenging.
Additionally, deep learning models can handle non-linear relationships and complex temporal patterns effectively, which traditional methods may struggle to capture \cite{ts2}. They can adapt to changing trends, seasonal variations, and irregular patterns in the data, making them versatile for a wide range of time series forecasting tasks.

In time series forecasting using deep learning, selecting an appropriate loss function is vital to guide the training process and optimize model performance. Commonly used loss functions for time series forecasting tasks include Mean Squared Error (MSE) and Mean Absolute Error (MAE), which quantify the discrepancy between predicted values and actual observations over the time series sequence. These loss functions aim to minimize the error between the predicted and true values, facilitating the model to learn the underlying patterns and trends efficiently. Additionally, custom loss functions like Quantile Loss can be employed for probabilistic forecasting, capturing uncertainty in the forecastings and enabling the model to provide prediction intervals \cite{ts3}. The choice of the loss function in time series forecasting using deep learning is crucial for enhancing the model's accuracy, capturing temporal dependencies effectively, and generating reliable forecasts for a wide range of applications including financial forecasting, demand forecasting, and anomaly detection.

\begin{table*}[t!]
\centering
\caption{Summary of time series forecasting methods and the used loss functions}
\label{forecasting}
\begin{tabular}{|p{3cm}|p{2.5cm}|p{4cm}|}
\hline
\textbf{Method} &  \textbf{Technique} &\textbf{Loss Function} \\ \hline
N-BEATS \cite{borovcnik2020n}& ANN &MSE \\ \hline
DeepAR \cite{salinas2020deepar}& RNN & Negative Log-Likelihood \\\hline
Lim et al.  \cite{lim2021time}& TR & MSE or Huber Loss \\ \hline
SCINet  \cite{Liu2022}     & CNN    &  L1     \\ \hline
Informer  \cite{Zhou2021}      &   TR   &   MSE   \\ \hline
FEDformer \cite{Zhou2022}    &  TR    &  MSE     \\ \hline
Salman et al.  \cite{Salman2024}     &  CNN, LSTM, TR    &  MSE     \\ \hline
RB-GRU-CNN   \cite{Khan2024}  &   CNN, GRU   &   Cross-Entropy    \\ \hline
Huang et al.    \cite{Huang2024}   &   Entropy features   &  Cross-Entropy     \\ \hline
King et al.     \cite{King2024}  &    CNN, RNN  &   Cross-Entropy    \\ \hline
\end{tabular}
\end{table*}
Recent advancements in time series forecasting have increasingly focused on various deep learning methods tailored for capturing the temporal patterns within data. For example,
N-BEATS (Neural Basis Expansion Analysis) is a powerful architecture specifically designed for univariate time series forecasting tasks, utilizing MSE for loss optimization while handling seasonal patterns and trends \cite{borovcnik2020n}.
 Furthermore, DeepAR, developed by Amazon, integrates recurrent layers with probabilistic forecasting, aiming to provide a distribution over forecasts and often employs negative log-likelihood loss during training \cite{salinas2020deepar}.
Recently, Transformers have gained popularity in natural language processing, and have also been adapted for time series forecasting. Their self-attention mechanism allows them to focus on relevant parts of the sequence, utilizing MSE or Huber Loss for optimization \cite{lim2021time}.



In the same context, SCINet introduced a novel recursive downsample-convolve-interact architecture that leverages the property of preserved temporal relations after downsampling. This architecture employs multiple convolutional filters to extract distinct yet valuable temporal features from downsampled sub-sequences at various resolutions. By aggregating these rich features, SCINet effectively models complex temporal dynamics inherent in time series forecasting tasks. While the L1 loss function is used  \cite{Liu2022}. In another work, Informer \cite{Zhou2021} addresses the challenges associated with long sequence time-series forecasting (LSTF) by proposing an efficient transformer model. Key features of Informer include a ProbSparse self-attention mechanism and techniques for self-attention distillation that enhance dependency alignment and performance. As loss function LSTF used MSE. Also, Based on the Transformer network, FEDformer combines Transformer architecture with seasonal-trend decomposition to capture both global trends and detailed structures in time series data. This method leverages the fact that many time series exhibit a sparse representation in the frequency domain, enhancing the transformer’s forecasting capabilities while maintaining linear complexity in relation to sequence length. The MSE loss function used \cite{Zhou2022}.

The study in \cite{Salman2024} investigated various hybrid models integrating convolutional neural networks (CNN), long short-term memory (LSTM), and transformer architectures for solar power forecasting. By evaluating all combinations of these models, particularly the CNN–LSTM–TF hybrid model, the research identifies that this model achieves the best forecasting accuracy by applying the MSE loss function.
In another research, the authors proposed a hybrid model combining GRU and CNN approaches to analyze ECG time series data, aimed at predicting survival probabilities for patients with various heart conditions. The RB-GRU-CNN model enhances the base GRU-CNN design by integrating a residual bias term, using a Cross-Entropy loss function \cite{Khan2024}.
The proposed information theory-based pipeline in \cite{Huang2024} addresses challenges in effectively recognizing patterns in neurological time series data. By employing various entropy methods, the pipeline captures essential statistical information while minimizing privacy concerns. While the Cross-Entropy loss function is applied.
In the same context, a CNN-RNN hybrid model for forecasting outlet fluid temperatures in Borehole Heat Exchanger (BHE) systems. By addressing challenges in feature extraction and nonlinear relationships, while using a Weighted Cross-Entropy loss function \cite{King2024}.


A summarization of the selected methods for time series forecasting and the loss functions used is provided in Table \ref{forecasting}.

\section{Computation Evaluation of Loss Function  }

In the field of computer vision and generative models, loss functions play a critical role in guiding the learning process and determining the quality of the generated results. Different architectural frameworks, such as Convolutional Neural Networks (CNNs), Vision Transformers (ViTs), and Generative Adversarial Networks (GANs), utilize tailored loss functions that align with their specific computational architectures and objectives. This section provide a description of loss functions across various architectures, as well as an evaluation of the computational aspects of these loss functions, assessing their efficiency, convergence properties, and impact on overall model performance. Understanding the computational implications is key for practitioners aiming to optimize training processes and resource utilization.
Also a  critically analyze the benefits and limitations of various loss functions, providing insights into their effectiveness in the computer vision tasks.

In the development of image generation models, the choice of loss function plays a pivotal role in shaping the model's performance and efficiency. This section delves into the computational aspects of various loss functions, with a focus on evaluating their computational efficiency, implementation complexity, and the ease with which gradients can be computed. Understanding these factors is crucial for optimizing model training and ensuring that the models can be deployed efficiently across different computational environments.

\subsubsection{Generative tasks}
In the realm of image generation, the choice of loss function can significantly impact model performance; thus, understanding the implications of each function is crucial. 

\textbf{Computational efficiency} is a primary concern, as complex loss functions such as Adversarial Loss and Diffusion Noise Prediction Loss may require substantial computational resources due to their reliance on additional networks or intricate calculations involved in noise modeling. In contrast, simpler functions like L1 Loss and Binary Cross-Entropy offer high efficiency with straightforward computations, allowing for faster training times across various applications.

\textbf{Implementation complexity} varies widely among these loss functions. Some, like Cross-Entropy and Categorical Cross-Entropy, are widely supported by deep learning frameworks and are easy to implement, making them favorable choices for many tasks. In contrast, loss functions that incorporate conditional aspects or complex overlap calculations, such as Dice Loss and Contrastive Loss, tend to involve more nuanced implementations. Additionally, loss functions that require integration of external architectures, such as Perceptual Loss and Variational Autoencoder Loss, present additional complexity due to the need to align with feature extractors or manage latent distributions.

\textbf{Gradient computation ease} When considering the ease of gradient computation, simpler losses such as Smooth L1, MSE, or standard Cross-Entropy yield straightforward derivatives that facilitate efficient optimization. Conversely, more complex losses like Jaccard Loss and Focal Loss can pose challenges in gradient computation, particularly when output distributions are close to decision boundaries or require weighting mechanisms to emphasize specific examples. This complexity can complicate convergence during training and may lead to instability if not handled carefully.

Overall, a balance must be struck between computational efficiency, implementation complexity, and gradient computation ease when selecting a loss function for image generation tasks. While simpler loss functions are easier to integrate and optimize, their effectiveness may vary depending on the specific use case and characteristics of the data. Conversely, more complex loss functions may provide superior performance in particular domains (e.g., dealing with class imbalances or enhancing perceptual quality) but may come with higher computational costs and implementation hurdles. 
In conclusion, the choice of loss function should align with the goals of the image generation task at hand, taking into account the trade-offs between computational demands, Ease of use, and the desired effectiveness in representing the nuances of the data.

\subsubsection{Discriminative tasks}
Evaluating the computational aspects of loss functions in discriminative tasks is essential for understanding model performance and optimizing the training process. Loss functions are crucial in measuring how well a model's predictions align with the actual target values. Here, we discuss the evaluation of computational aspects of loss functions on discriminative tasks.

\textbf{Computational Efficiency}
When examining computational efficiency, loss functions exhibit varying performance profiles that are essential to consider in practical applications. Functions like Euclidean Loss, Mean Squared Error (MSE), and Binary Cross-Entropy generally excel in this regard, as they rely on simple arithmetic operations and can leverage vectorized implementations in deep learning frameworks. They allow for rapid calculations, especially useful in models with large datasets. In contrast, loss functions such as Attention-Target Loss and Gaussian Wasserstein Distance are more computationally intensive due to the additional complexities introduced by attention mechanisms and optimal transport calculations. These can significantly slow down the training process, particularly with large input sizes or when many computations are needed.

Furthermore, losses that use pairwise comparisons, such as Ranking Loss and Contrastive Loss, can be computationally demanding depending on the number of examples paired together. As the dataset size increases, the cost associated with calculating these comparisons can become substantial, impacting overall training efficiency. Conversely, loss functions tailored for segmentation tasks, like Intersection over Union (IoU), also experience high computational demands due to the requirement for area calculations between predicted and ground-truth regions, further emphasizing the computational burden associated with certain losses.

\begin{table*}[ht]
    \centering
    \caption{Advantages and Disadvantages of Common Loss Functions}
    \label{adv}
    \begin{tabular}{|p{3cm}|p{6cm}|p{7cm}|}
        \hline
        \textbf{Loss Function} & \textbf{Advantages} & \textbf{Disadvantages} \\
        \hline
        Cross-entropy loss & Effective in multi-class classification; Provides probabilistic interpretations; Easy to implement. & Sensitive to class imbalance; Can lead to overconfidence in predictions. \\
        \hline
        Binary Cross-Entropy Loss & Simple and efficient for binary classification; Provides clear probabilistic outputs. & Assumes outputs are independent; may not perform well with correlated data. \\
        \hline
        Weighted Cross-Entropy Loss & Helps mitigate class imbalance; Adjusts the importance of classes. & Requires careful tuning of weights; Increased computational overhead. \\
        \hline
        P-w Softmax with Cross-Entropy Loss & Effective for segmentation tasks; Maintains spatial context. & High computational and memory requirements; Complex gradients. \\
        \hline
        L1 Loss & Robust to outliers; Minimizes extreme values' impact. & Can lead to less smooth convergence during optimization. \\
        \hline
        Dice Loss & Well-suited for imbalanced datasets; Focuses on overlap rather than pixel-wise accuracy. & Gradients can be complex near boundaries and harder to optimize. \\
        \hline
        Focal Loss & Addresses class imbalance by down-weighting easy examples. & More computationally demanding than standard cross-entropy; Requires tuning of the focusing parameter. \\
        \hline
        Jaccard Loss & Directly optimizes overlap measures; Ideal for segmentation tasks. & Computationally intensive; Complex gradient calculations. \\
        \hline
        Adversarial Loss & Enhances realism in generated images; Models distribution well. & Requires simultaneous training of generator and discriminator; Can lead to instability. \\
        \hline
        Categorical Cross-Entropy Loss & Works well for multi-class problems; Provides probabilistic outputs. & Class imbalance issues observed; Similar to standard cross-entropy. \\
        \hline
        Contrastive Loss & Effective for learning similarities and differences between pairs. & Requires careful curation of pairs; Increases implementation complexity. \\
        \hline
        Diffusion Noise Prediction Loss & Useful in modern generative models; Handles noise explicitly. & High computational burden; Model complexities can complicate training. \\
        \hline
        Perceptual Loss & Captures perceptual differences; Leads to visually appealing results. & Computationally expensive due to feature extraction; May require significant model adjustments. \\
        \hline
        Variational Autoencoder (VAE) Loss & Provides a probabilistic framework for generative models. & Balancing reconstruction and KL divergence can be tricky; May need careful tuning of parameters. \\
        \hline
    \end{tabular}
    
\end{table*}

\textbf{Implementation Complexity}
Implementation complexity is another critical consideration that influences the choice of loss functions. Losses like Cross-Entropy and Binary Cross-Entropy are well-documented and commonly implemented in most deep-learning libraries, making them straightforward to integrate into various models. This ease of use is a stark contrast to functions such as Attention-Target Loss and Gaussian Wasserstein Distance, which require more intricate architectural design and an understanding of model dynamics for proper integration. The latter can make the implementation of these loss functions challenging and may necessitate deeper expertise.

Moreover, loss functions requiring additional parameters for tuning, such as Focal Loss (which addresses class imbalance by adjusting the focus on hard examples), add another layer of complexity as they necessitate careful tuning to achieve optimal performance. Similarly, IoU Loss and its efficient variants need meticulous implementation to ensure accurate calculation of overlaps, which can introduce potential sources of error if not executed correctly.

\textbf{Ease of Gradient Computation}
Ease of gradient computation is also a fundamental factor in the choice of loss functions. Functions like MSE and Cross-Entropy provide straightforward gradient calculations that simplify the optimization process, contributing to effective and rapid convergence during training. In contrast, loss functions that involve non-linear calculations, such as Contrastive Loss and Ranking Loss, can complicate gradient computations. These functions rely on the relationships between data points rather than absolute values, making their derivatives more complex and potentially leading to slower convergence rates if not managed properly.

Furthermore, gradient calculations for loss functions like IoU may prove challenging, particularly when the outputs approach critical thresholds (as in segmentation scenarios). Similarly, losses involving more advanced metrics, such as Mahalanobis Distance, can produce intricate gradients that require careful handling to avoid difficulties during the training process.

In summary, the choice of loss functions must be informed by their computational efficiency, implementation complexity, and ease of gradient computation. Understanding these practical considerations allows practitioners to select loss functions that maximize performance while aligning with the specific requirements of their models and datasets in image classification, object detection, and action recognition tasks. Balancing these aspects is crucial for ensuring effective model training and deployment.

\section{Benefits, Challenges and Future Directions}

\subsection{Benefits and Limitations}

Many loss functions exhibit noteworthy advantages that make them suitable for various tasks in image generation. Cross-Entropy Loss and its variants, such as Binary Cross-Entropy and Categorical Cross-Entropy, offer robust performance in classification tasks, providing probabilistic interpretations that make them easy to implement. The Weighted Cross-Entropy extends this advantage by allowing for effective handling of class imbalances, making it ideal for datasets where certain classes are underrepresented. For segmentation tasks, Pixel-wise Softmax with Cross-Entropy Loss maintains spatial context and excels at guiding models to optimize pixel-level accuracy.

Loss functions like L1 Loss and Dice Loss are particularly advantageous in scenarios involving outliers or imbalanced datasets. L1 Loss is robust to extreme values, minimizing their influence on model training, while Dice Loss emphasizes overlap between predicted and ground truth regions, providing better performance in situations with class imbalance. Focal Loss is advantageous in handling challenging class distributions by down-weighting easy examples, thus focusing more on hard-to-classify data. The Adversarial Loss enhances the realism of generated images by training models to mimic data distributions effectively. Moreover, Perceptual Loss captures perceptual differences, leading to visually appealing results, while Variational Autoencoder Loss offers a probabilistic framework that promotes generation diversity.

\textbf{Disadvantages of Loss Functions}
Despite their advantages, each loss function comes with its own set of disadvantages that can limit their applicability or efficiency. Functions such as Cross-Entropy Loss, Binary Cross-Entropy, and Weighted Cross-Entropy can be sensitive to class imbalances and may lead to overconfident predictions if not properly calibrated. Similarly, while Pixel-wise Softmax with Cross-Entropy provides detailed segmentation, it requires significant computational resources, which may not be feasible for large images or datasets.

Loss functions like L1 Loss and Dice Loss can present challenges in convergence; L1 Loss may lead to less smooth optimization paths, while Dice Loss can exhibit complex gradients near class boundaries, complicating the training process. Focal Loss has increased computational demands and requires careful tuning of its focusing parameter, potentially complicating implementation. Jaccard Loss and its computationally intensive nature can hinder performance, especially in real-time applications.

When utilizing Adversarial Loss, stability during training can be a significant concern since it necessitates concurrent training of both generators and discriminators, which can complicate the training dynamics. Furthermore, while Perceptual Loss captures important visual characteristics, it is computationally expensive due to its reliance on deep feature extraction networks. Lastly, Variational Autoencoder (VAE) Loss, while powerful, can be difficult to balance effectively between reconstruction error and KL divergence, requiring careful parameter tuning to achieve optimal results.

The selection of loss functions in image generation must consider both their advantages and disadvantages, aligning them with the specific requirements of the task and the characteristics of the data at hand. A summarization of some selected loss functions in terms of advantages and disadvantages is provided in Table \ref{adv}.

\subsection{Challenges}

\begin{figure}
    
\begin{tikzpicture}
[decoration={start radius=0.5cm, end radius=.5cm, amplitude=3mm, angle=30}]

\colorlet{outputcolor}{blue!50}
\colorlet{outputcolor1}{blue}
\colorlet{robustnesscolor}{green!70}
\colorlet{outliercolor}{orange!80}

\begin{scope}[mindmap,
every node/.style={concept, circular drop shadow, minimum size=0pt, execute at begin node=\hskip0pt, font=\bfseries},
root concept/.append style={
    concept color=black, fill=white, line width=1.5ex,,minimum size=2cm,scale=0.9, text=black, font=\normalsize\scshape\bfseries,},
level 1 concept/.append style={font=\bfseries},
text=white,
output/.style={concept color=outputcolor},
output1/.style={concept color=outputcolor1},
robustness/.style={concept color=robustnesscolor},
outlier/.style={concept color=outliercolor},
grow cyclic,
level 1/.append style={level distance=4.2cm, sibling angle=75},
level 2/.append style={level distance=3cm, sibling angle=75}]
\node [root concept] {Challenges}[rotate=202.5] 
    child [output1] { node {Nature of Outputs} 
        child [output]  { node {Categorical Outputs} }
        child [output] { node {Continuous Outputs} }
        child [output] { node {Structured Outputs} }
        child [output] { node {Multi-modal Outputs} } 
    }
    child [robustness] { node {Robustness and Generalization} }
    child [outlier] { node {Sensitivity to Outliers or Noisy Data} };
\end{scope}

\end{tikzpicture}
  \caption{{Challenges related to the loss function. }}
\label{chall}
\end{figure}
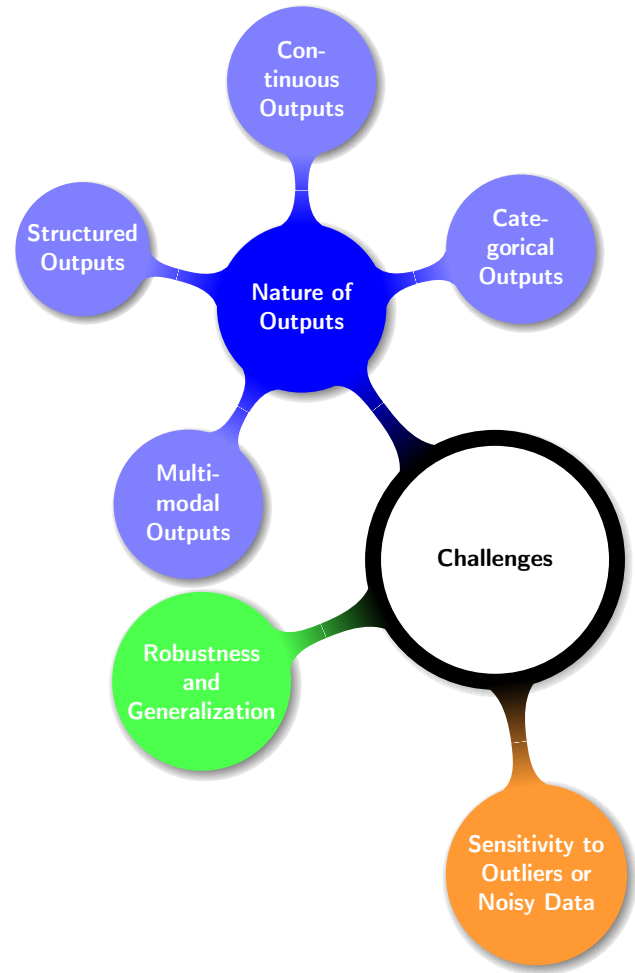

Choosing the right loss function for a deep learning task is a critical determinant of the performance and success of a model. Each task in deep learning, whether it is regression, classification, or any other, requires a loss function that is aligned with the specific objectives and characteristics of the task. However, several challenges arise in this selection process. These include the appropriate reflection of the task's goals, ensuring the loss function exhibits robustness and generalization across varied data distributions, and addressing sensitivity to outliers or noisy data, which can significantly affect model performance. These challenges illustrated in Figure \ref{chall} will be discussed in details in this section.

\textbf{The appropriate reflection (of the nature of output):}

One major challenge is ensuring that the loss function appropriately reflects the nature of the output. For example, in regression tasks where the goal is to predict continuous values, Mean Squared Error (MSE) is commonly used, whereas for classification tasks, Cross-Entropy Loss is preferred. However, the situation becomes complex when dealing with multi-label classification or semantic segmentation, where standard loss functions might not capture the nuances of class imbalance or intricate relationships between outputs, requiring custom loss formulations or weighting schemes.

In image segmentation, the primary challenge is handling class imbalances where certain classes may dominate the image, leading traditional loss functions like Cross-Entropy to perform poorly. To address this, variations such as Weighted Cross-Entropy or Dice Loss are often used to ensure that each class is appropriately considered during the training process. Dice Loss, in particular, is effective because it directly maximizes the overlap between predicted and true segments, which is crucial for applications like medical imaging where precise boundaries are critical.

\underline{Nature of Outputs:}
The nature of the outputs varies widely across different tasks—ranging from discrete class labels in classification tasks to continuous pixel values in image generation. For each of these different types of outputs, the choice of loss function must be tailored to effectively capture the nuances of the task at hand.

\underline{Categorical Outputs:}
In classification tasks, a common loss function is Cross-Entropy Loss, which is designed to handle categorical outputs effectively. However, when dealing with imbalanced datasets where some classes appear much less frequently than others, traditional cross-entropy might not perform well. Solutions like using class weights can help, but they may still fall short in capturing the true relationships among classes \cite{Cui2019}.

\underline{Continuous Outputs:}
In regression and image generation tasks, Mean Squared Error (MSE) is often used. However, MSE tends to penalize larger errors disproportionately, which can be problematic in cases where outliers exist in the data. Alternatives such as Huber Loss or Log-Cosh Loss can provide a more balanced approach that mitigates this issue while still reflecting the nature of continuous predictions \cite{Huber1964}.

\underline{Structured Outputs:}
Tasks such as semantic segmentation or object detection require the output to be structured (i.e., pixel-wise predictions). Loss functions need to account for the spatial dependencies and relationships among predicted pixels. Dice Loss or Intersection over Union (IoU) Loss has proven effective in this context as it focuses on the overlap between predicted and true segmentations, directly addressing the characteristics of the output \cite{Milletari2016}.

\underline{Multi-modal Outputs:}
In tasks involving multiple types of outputs, such as text-to-image generation, the loss function must bridge diverse modalities. Here, composite loss functions that incorporate perceptual loss, such as the content loss derived from feature extractors, are used to ensure that the generated images align well with the underlying textual descriptions \cite{Ramesh2021}.

\underline{Robustness and Generalization:}
A critical consideration is ensuring that the loss function remains robust to various forms of noise and outliers within the data. The loss function should capture not only the most common cases but also accommodate variability and ensure generalization to unseen data. Techniques such as data augmentation and the use of regularized loss functions can enhance this aspect \cite{Shorten2019}.

In summary, aligning the loss function with the characteristics of the output is a significant challenge that influences model performance across various computer vision tasks. The selection of an appropriate loss function is crucial for effectively capturing the underlying relationships in the data, accommodating imbalances, and ensuring robustness to noise. Continuous research is needed to develop more sophisticated loss functions that reflect the complexities and variabilities of real-world data.

\textbf{Sensitivity to outliers or noisy data:}

Another challenge arises from the sensitivity of loss functions to outliers or noisy data. Some loss functions, like Mean Squared Error, are highly sensitive to outliers, which can skew the training process and result in poor generalization. In such cases, alternatives like Huber Loss or quantile-based losses may be more robust and yield better performance. Moreover, in tasks involving structured or sequential data, such as language modeling or time series forecasting, the choice of loss function can significantly influence the convergence and the quality of predictions \cite{ts2}. The challenge is to balance computational efficiency with the need for capturing temporal dependencies and maintaining stability throughout the learning process.

Loss functions that calculate errors directly, such as Mean Squared Error (MSE), can be significantly affected by outliers. Since MSE squares the error, a single large error can disproportionately influence the overall loss, leading to model parameters that do not generalize well to the majority of the data. In contrast, loss functions like Huber Loss combine the advantages of MSE and Mean Absolute Error (MAE), being less sensitive to outliers by applying a quadratic loss for small errors and a linear loss for larger ones \cite{Huber1964}.

In semantic segmentation, mislabeled pixels (noisy data) can lead to poor model performance. If the loss function is sensitive to these misclassifications, it can skew the learning process, causing the model to fit to noise rather than learning the true distribution of data. Techniques like label smoothing can help mitigate this issue by softening class labels \cite{Szegedy2016}.

Adversarial models such as GANs can also display sensitivity to noisy data. If the discriminator learns from noisy samples, it might lead the generator to produce artifacts. Employing regularization techniques or using loss functions robust to noise can enhance performance. For example, Wasserstein Loss is known to be more stable in the presence of outliers \cite{Arjovsky2017}.

In these multi-modal tasks, noise in the input data can disrupt the alignment between text and visual features or between audio and images. Loss functions must be designed to withstand variations and noise in inputs, often requiring the integration of robust training strategies, such as curriculum learning, where the model is gradually exposed to more complex examples to build its robustness to noise \cite{Bengio2009}.

To reduce the impact of outliers and noise, practitioners can employ robust loss functions, outlier detection methods, and data preprocessing techniques. For instance, using data augmentation can help models generalize better by exposing them to a wider range of variations, thereby reducing sensitivity to noise in the core data \cite{Shorten2019}.

In conclusion, understanding the sensitivity of loss functions to outliers and noisy data is critical for developing robust machine learning models in computer vision. By selecting appropriate loss functions and incorporating techniques to minimize the effects of noise, one can enhance model performance and generalizability. 

\begin{figure*}
\centering
\begin{tikzpicture}
[decoration={start radius=1cm, end radius=.5cm, amplitude=1mm, angle=30}]

\colorlet{adaptivecolor}{blue!50}
\colorlet{modalcolor}{green!70}
\colorlet{hierarchicalcolor}{orange!80}
\colorlet{explainablecolor}{violet!70}
\colorlet{metalearningcolor}{red!60}
\colorlet{efficiencycolor}{teal!80}
\colorlet{contextcolor}{purple!50}

\begin{scope}[mindmap,
every node/.style={concept, circular drop shadow, minimum size=0pt, execute at begin node=\hskip0pt, font=\bfseries},
root concept/.append style={
    concept color=black, fill=white, line width=1.5ex,scale=0.9, text=black, font=\Large\scshape\bfseries,},
level 1 concept/.append style={font=\bfseries},
text=white,
adaptive/.style={concept color=adaptivecolor},
modal/.style={concept color=modalcolor},
hierarchical/.style={concept color=hierarchicalcolor},
explainable/.style={concept color=explainablecolor},
metalearning/.style={concept color=metalearningcolor},
efficiency/.style={concept color=efficiencycolor},
context/.style={concept color=contextcolor},
grow cyclic,
level 1/.append style={level distance=4.5cm, sibling angle=45},
level 2/.append style={level distance=3cm, sibling angle=45}]
\node [root concept] {Future Directions}[rotate=20.5] 
    child [adaptive] { node {Adaptive Loss Functions} }
    child [modal] { node {Multi-Modal Loss Functions} }
    child [hierarchical] { node {Hierarchical and Structured Loss Functions} }
    child [explainable] { node {Explainable Loss Functions} }
    child [metalearning] { node {Meta-Learning Losses} }
    child [efficiency] { node {Efficiency in Scale} }
    child [context] { node {Context-Aware Loss Functions} };
\end{scope}

\end{tikzpicture}
  \caption{{Future directions of loss functions.}}
\label{future}
\end{figure*}
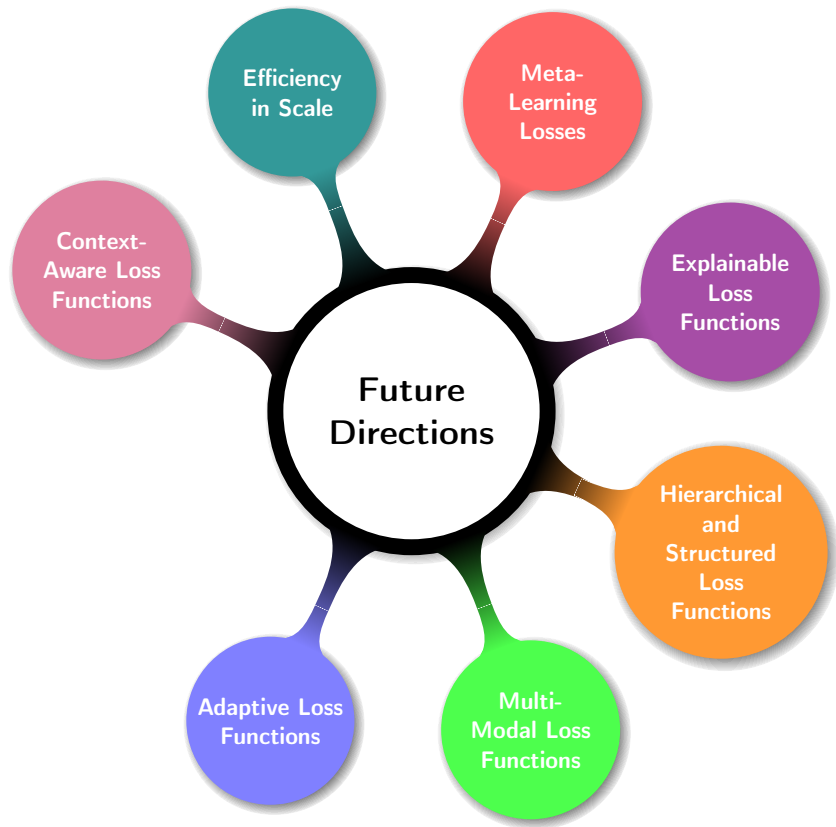

\textbf{Interpretability and scalability:}

Additionally, the interpretability and scalability of the loss function pose challenges, especially in complex scenarios like adversarial training or reinforcement learning, where the loss must not only drive the model towards better predictions but also ensure stability and convergence in a dynamic environment. In adversarial networks, for example, the opposing objectives of generator and discriminator networks require carefully designed loss functions to ensure they evolve in harmony. Similarly, in reinforcement learning, the reward-based loss functions need to be meticulously crafted to reflect the true value of actions in terms of future benefits. Customizing and tuning loss functions according to these diverse requirements often involves significant experimentation and domain knowledge, presenting an ongoing challenge for practitioners working on deep learning models.

For example, for Semantic Segmentation, the choice of loss function directly impacts the model's performance in segmenting images into meaningful components. Commonly used loss functions such as Cross-Entropy Loss and Dice Loss must align with human intuition, reflecting improvements in segmentation quality. As the number of classes or the resolution of images increases, scalability becomes a notable challenge, necessitating efficient handling of computational demands. Techniques such as class weighting or adaptive sampling can help mitigate these challenges \cite{Milletari2016}.

In image generation tasks, models like Generative Adversarial Networks (GANs) employ adversarial loss, which poses interpretability challenges. This loss measures the generator's capacity to fool the discriminator rather than providing a direct assessment of image quality. Scalability is also critical, especially with growing datasets and model architectures. Implementing strategies such as the progressive growth of GANs enables effective management of these computational demands \cite{Karras2018}.

The integration of text and visual data in generation tasks requires loss functions that capture not only visual realism but also semantic alignment with the input text. This complexity often necessitates composite losses that include perceptual and matching-aware metrics. As the complexity of text prompts grows, so does the need for scalability in loss functions. Modern architectures, such as transformers, have shown promise in effectively scaling these tasks \cite{Ramesh2021}.

This emerging area presents unique interpretability challenges, as the loss functions must account for both auditory features and visual coherence. Balancing these components is crucial yet complex. Moreover, the process of managing high-dimensional audio data alongside high-resolution images requires robust data processing techniques to ensure scalability without compromising performance \cite{Gao2021}.

In summary, addressing the challenges of interpretability and scalability in loss functions for computer vision tasks calls for the development of hybrid models and innovative loss functions that incorporate multi-modal metrics while maintaining computational efficiency. Ongoing research continues to explore new pathways to enhance the effectiveness and usability of these models.


\subsection{Future directions}

The future directions for loss functions in deep learning across various tasks can focus on several key areas that aim to enhance robustness, interpretability, efficiency, and adaptability. As represented in Figure \ref{future} some future directions that describe different solutions will be presented in this section.

\textbf{Adaptive Loss Functions}: Developing loss functions that can adapt dynamically based on the data distribution or the presence of outliers. For instance, creating loss functions that automatically adjust their sensitivity to noisy data could lead to improved performance in real-world applications.

\textbf{Robustness to Adversarial Attacks}: As adversarial examples are a significant concern in deep learning, future loss functions could be designed to mitigate the impact of such inputs through enhanced focus on adversarial training or by explicitly incorporating adversarial loss components \cite{Goodfellow2014}.

\textbf{Unified Frameworks for Multi-Modal Learning}: As models increasingly work with multiple modalities (e.g., text, image, audio), future loss functions could be designed to handle and integrate these different types of data more effectively. This may involve composite loss functions that capture the relationships and dependencies between modalities, aiding in tasks such as text-to-image generation and audio-to-image translation \cite{Baltrusaitis2019}.

\textbf{Loss Functions for Hierarchical Outputs}: For tasks that involve structured outputs, such as semantic segmentation and image generation, future research may focus on developing hierarchical loss functions that account for the relationships between different parts of the output. These functions could help in maintaining spatial coherence and context within predictions \cite{Zhang2018}.

\textbf{Interpretable Loss Structures}: As interpretability becomes increasingly important in machine learning, the design of loss functions that provide transparent and understandable metrics of performance could enhance trust in models. Future loss functions may incorporate components that clearly illustrate what aspects of the prediction are being penalized or rewarded \cite{Ribeiro2016}.


\textbf{Meta-Learning Approaches}: Exploring the integration of meta-learning frameworks with loss functions that can learn to adjust themselves based on the training context and data characteristics. This could enable models to perform better across diverse datasets and tasks, enhancing generalizability \cite{Finn2017}.

\textbf{Scalable Loss Functions}: In light of the growing size of datasets and model architectures, future loss functions could be optimized for computational efficiency. Research may explore ways to reduce the complexity of loss calculations while maintaining their effectiveness, potentially through approximate methods or hierarchical loss architectures \cite{Brock2018}.

\textbf{Incorporating Contextual Information}: Future loss functions could be developed to utilize contextual information within the data, enhancing their sensitivity to important features that would improve model learning without overfitting \cite{Zhou2019}.

In conclusion, the future directions for loss functions in deep learning will likely emphasize robustness, adaptability, efficiency, interpretability, and multi-modality to better handle the complexities of real-world tasks. Addressing these areas can lead to more effective and reliable models across various domains. Also, by developing innovative loss functions that align more closely with the intricacies of real-world data and tasks, researchers can improve model performance and reliability. This work will be crucial to addressing ongoing challenges in deep learning and expanding its applicability across diverse domains. These advancements in loss function design will not only refine model training but also contribute to the broader goal of creating explainable AI, achieving higher efficiency in scaling, and effectively managing the complexities inherent in multi-modal learning environments.


\section{Conclusion}
This paper has presented a comprehensive overview of loss functions used in various deep learning applications, with a strong emphasis on computer vision tasks. We have explored the diverse landscape of loss functions, ranging from standard metrics to specialized variants tailored to specific challenges. The historical overview highlighted the evolution of these functions alongside advancements in deep learning architectures. While effective loss functions are crucial for successful model training, the inherent trade-offs between computational efficiency, robustness to noise, and the ability to capture complex relationships in the data necessitate ongoing research. Future directions include developing adaptive, interpretable, and scalable loss functions, particularly for addressing multi-modal data and the challenges posed by adversarial attacks. By carefully considering the characteristics of the data and task objectives, researchers can leverage the insights presented here to design more effective and robust deep-learning models.

\section*{Acknowledgments}

This research work was made by research grant support (G00004666) from College of Information Technology (CIT), United Arab Emirates University (UAEU).

\section*{Appendix: Definitions and Formulations}

\section*{Mean Squared Error (MSE)}
MSE is a common loss function for regression tasks, assessing the average squared difference between predicted and actual values:
\[
L(y, \hat{y}) = \frac{1}{N} \sum_{i=1}^{N} (y_i - \hat{y}_i)^2
\]
where \(N\) is the number of samples, \(y\) represents the true values, and \(\hat{y}\) denotes the predicted values. MSE penalizes larger errors more than smaller ones.

\section*{Mean Absolute Error (MAE)}
MAE is another regression loss function that measures the average absolute difference between predicted and true values:
\[
L(y, \hat{y}) = \frac{1}{N} \sum_{i=1}^{N} |y_i - \hat{y}_i|
\]
MAE is less sensitive to outliers compared to MSE.

\section*{Cross-Entropy Loss}
Cross-Entropy Loss measures the performance of a classification model whose output is a probability value between 0 and 1. It evaluates the difference between two probability distributions — the true labels and the predicted probabilities. 
\[
L(y, \hat{y}) = - \sum_{i=1}^{N} y_i \log(\hat{y}_i)
\]
where \(y_i\) is the ground truth label (0 or 1) and \(\hat{y}_i\) is the predicted probability for class \(i\). This loss is used extensively for classification tasks.

\section*{Binary Cross-Entropy Loss}
Binary Cross-Entropy Loss is used for binary classification tasks. It assesses the difference between the true binary labels and the predicted probabilities:
\[
L(y, \hat{y}) = - \left( y \log(\hat{y}) + (1 - y) \log(1 - \hat{y}) \right)
\]
where \(y\) is the true binary label and \(\hat{y}\) is the predicted probability.

\section*{Gaussian Wasserstein Distance}
The Gaussian Wasserstein Distance between two bounding box distributions can be calculated as follows:
\[
W(B_{pred}, B_{gt}) = \sqrt{(x - x')^2 + (y - y')^2 + (w - w')^2 + (h - h')^2}
\]
Here, the distance accounts for differences in position, width, and height, pooling information about the bounding boxes into a single measure.

\section*{Hinge Loss}
Hinge Loss is primarily used for training classifiers such as Support Vector Machines (SVMs). It is defined as:
\[
L(y, \hat{y}) = \max(0, 1 - y \cdot \hat{y})
\]
where \(y \in \{-1, 1\}\) is the true label, and \(\hat{y}\) is the predicted output. The loss is zero if the prediction is correct and within a margin of 1, otherwise, it increases linearly.

\section*{Focal Loss}
Focal Loss is designed to address class imbalance by putting more focus (i.e., loss) on hard-to-classify examples. It modifies the standard Cross-Entropy Loss by adding a modulating factor:
\[
L(y, \hat{y}) = - \alpha (1 - \hat{y})^\gamma \log(\hat{y})
\]
where \(\alpha\) is a scaling factor and \(\gamma\) is a focusing parameter. When \(\gamma = 0\), Focal Loss corresponds to Cross-Entropy Loss.

\section*{Smooth L1 Loss (Huber Loss)}
Smooth L1 Loss, also known as Huber Loss, is a combination of Mean Squared Error (MSE) and Mean Absolute Error (MAE) that is less sensitive to outliers in data:
\[
L(y, \hat{y}) = \begin{cases} 
\frac{1}{2} (y - \hat{y})^2 & \text{if } |y - \hat{y}| < 1 \\
|y - \hat{y}| - \frac{1}{2} & \text{otherwise}
\end{cases}
\]
where \(y\) is the true value and \(\hat{y}\) is the predicted value. This loss is particularly popular in object detection tasks.

\section*{Intersection over Union (IoU) Loss}
IoU Loss evaluates the overlap between the predicted bounding box and the ground truth bounding box in object detection tasks. It is defined as:
\[
\text{IoU} = \frac{\text{Area of Overlap}}{\text{Area of Union}}
\]
This measure ensures that the predicted bounding box tightly fits the ground truth.

\section*{Dice Loss}
Dice Loss is often used in image segmentation. It measures the overlap between the predicted and the ground truth masks:
\[
L(y, \hat{y}) = 1 - \frac{2 \sum_{i}^{N} y_i \hat{y}_i}{\sum_{i}^{N} y_i + \sum_{i}^{N} \hat{y}_i}
\]
where \(y\) is the ground truth mask and \(\hat{y}\) is the predicted mask. This metric emphasizes the overlap between predictions and the ground truth.

\section*{Jaccard Loss}
Jaccard Loss, also known as the Intersection over Union (IoU) Loss, is a popular choice for segmentation tasks:
\[
L(y, \hat{y}) = 1 - \frac{\sum_{i}^{N} y_i \hat{y}_i}{\sum_{i}^{N} y_i + \sum_{i}^{N} \hat{y}_i - \sum_{i}^{N} y_i \hat{y}_i}
\]
This loss function enhances segmentation performance by focusing on the intersection and union of the predicted and ground truth masks.

\section*{Perceptual Loss}
Perceptual Loss, also known as Feature Loss, is used in image generation tasks. It compares high-level difference between images using features extracted from a pre-trained network, such as VGG:
\[
L(y, \hat{y}) = \sum_{i} \left\| \phi_i(y) - \phi_i(\hat{y}) \right\|_2^2
\]
where \(\phi_i\) represents the feature maps extracted from layer \(i\) of the network.

\section*{Adversarial Loss (GANs)}
Adversarial Loss is employed in Generative Adversarial Networks (GANs) to fool the discriminator by generating realistic data. The loss for the discriminator \(D\) and generator \(G\) is defined as:
\[
L_D = - \mathbb{E} [\log D(x)] - \mathbb{E} [\log (1 - D(G(z)))]
\]
\[
L_G = - \mathbb{E} [\log D(G(z))]
\]
where \(x\) is a real sample, \(z\) is a noise vector, and \(G(z)\) is the generated sample.

\section*{Feature Matching Loss}
Feature Matching Loss, used in GANs, minimizes the difference between features produced by the discriminator for real and generated images:
\[
L_{FM}(y, \hat{y}) = \sum_{i=1}^{L} \left\| D^i(y) - D^i(\hat{y}) \right\|_1
\]
where \(D^i\) denotes the feature maps at the \(i\)-th layer of the discriminator.

\section*{Content Loss (Style Transfer)}
Content Loss in style transfer measures the difference in content between the generated image and the content image using feature maps from a pre-trained network:
\[
L_{content} = \sum_{i} \left\| \phi_i(y_{content}) - \phi_i(\hat{y}) \right\|_2^2
\]

\section*{Style Loss (Style Transfer)}
Style Loss assesses the difference in style between the generated image and the style image by comparing their Gram matrices:
\[
L_{style} = \sum_{i} \left\| G(\phi_i(y_{style})) - G(\phi_i(\hat{y})) \right\|_2^2
\]
where \(G(\phi_i)\) denotes the Gram matrix of the feature maps from layer \(i\).







\section*{Kullback-Leibler (KL) divergence}
The Kullback-Leibler (KL) divergence loss function is a measure of how one probability distribution diverges from a second, expected probability distribution. In machine learning and statistics, it is often used to quantify how well the predicted probability distribution (P) approximates the true distribution (Q).
The KL divergence from distribution ( Q ) to distribution ( P ) for discrete distributions is defined as:
\[
D_{\text{KL}}(Q \,||\, P) = \sum_{i} Q(i) \log\left(\frac{Q(i)}{P(i)}\right)
\]
For continuous distributions, it is defined as:
\[
D_{\text{KL}}(Q \,||\, P) = \int_{-\infty}^{+\infty} q(x) \log\left(\frac{q(x)}{p(x)}\right) \, dx
\]
where:
(Q) is the true distribution you want to approximate.
(P) represents the predicted distribution.

\section*{Content Loss}
The content loss function ensures that the generated image retains the essential content features of the original image. In neural style transfer, it's typically calculated using a pre-trained CNN, like VGG19.
\[
L_{\text{content}}(I_c, I_g, l) = \frac{1}{2} \sum_{i,j} \left( \Phi_l(I_g){i,j} - \Phi_l(I_c){i,j} \right)^2
\]
where:
$(I_c)$ is the content image.
$(I_g)$ is the generated image.
$(\Phi_l)$ is the feature map at layer (l).

\section*{Ranking Loss}

Ranking Loss measures how well the predicted rankings of items reflect the true rankings. It penalizes the model when items that should be ranked higher are ranked lower than items that should be ranked lower.

For a pair of items ( (i, j) ) where item ( i ) should rank higher than item ( j ), the ranking loss can be defined as:
\[
L_{\text{ranking}} = \sum_{(i,j) \in R} \max(0, s_j - s_i + \Delta)
\]
where:
$R$ is the set of all relevant item pairs where ( i ) should rank higher than ( j ). $s_i$ and $s_i$ are the predicted scores for items $i$ and $j$, respectively. $\Delta$ is a margin that defines the minimum difference required between the scores for the correct ranking.
\section*{Weighted cross-entropy loss}
Weighted cross-entropy loss is an extension of the standard cross-entropy loss function that assigns different weights to classes. This is particularly useful in scenarios where there's an imbalance in the class distribution, helping to mitigate the effects of class imbalance in binary or multi-class classification problems.

Weighted cross-entropy introduces a weight $w_c$ for each class $c$:
\[
L_{\text{WCE}} = -\frac{1}{N} \sum_{i=1}^{N} \sum_{c=1}^{C} w_c \cdot y_{i,c} \cdot \log(p_{i,c})
\]
Where:
$w_c$ is the weight associated with class $c$.

\section*{Reconstruction Loss}
Reconstruction loss quantifies the difference between the original image and the image produced by the model. The goal is to minimize this loss during training, ensuring that the generated images closely resemble the input images.

One commonly used form of reconstruction loss is the Mean Squared Error (MSE), which can be expressed mathematically as:

\[
L_{rec} = \frac{1}{N} \sum_{i=1}^{N} || x_i - \hat{x}_i ||^2
\]
Where:
$L_{rec}$: The reconstruction loss.
$N$: The number of images in the dataset or batch.
$x_i$: The original image.
$\hat{x}_i$: The reconstructed (or generated) image produced by the model.
$|| \cdot ||$: The Euclidean norm (L2 norm) that measures the distance between the two images.

\section*{Mahalanobis distance}
Mahalanobis distance is a measure of the distance between a point and a distribution, which accounts for the correlations of the data set and scales the distances based on the variance of the data in each dimension. In the context of loss functions, Mahalanobis distance is often used in various machine learning applications, such as clustering, anomaly detection, and metric learning.

The Mahalanobis distance between a point $x$ and a mean $\mu$ of a distribution with covariance matrix $S$ is defined as:

\[
D_M(x, \mu) = \sqrt{(x - \mu)^T S^{-1} (x - \mu)}
\]
Where:
$x$ is the data point.
$\mu$ is the mean of the distribution.
$S$ is the covariance matrix of the data.
$D_M(x, \mu)$ is the Mahalanobis distance from point $x$ to the mean $\mu$.

\section*{Logistic regression Loss}
Logistic regression loss function, commonly referred to as binary cross-entropy loss, is used in binary classification tasks. It quantifies the difference between the predicted probabilities of the positive class and the actual binary outcomes.

The logistic regression loss function (binary cross-entropy loss) is defined as:

\[
L(y, p) = -\frac{1}{N} \sum_{i=1}^{N} \left( y_i \log(p_i) + (1 - y_i) \log(1 - p_i) \right)
\]
Where:
$N$ is the number of samples.
$y_i$ is the true binary label for the $i-th$ instance (1 for positive class, 0 for negative class).
$p_i$ is the predicted probability of the positive class for the $i^-th$ instance.

\section*{Categorical Loss}
Categorical loss, specifically referred to as categorical cross-entropy loss, is a commonly used loss function for multi-class classification tasks. It measures the dissimilarity between the true distribution (one-hot encoded labels) and the predicted distribution (probabilities for each class) produced by a model.

In a multi-class classification problem, the categorical cross-entropy loss is defined as:

\[
L(y, p) = -\sum_{i=1}^{N} \sum_{c=1}^{C} y_{i,c} \log(p_{i,c})
\]
Where:
$N$ is the number of samples.
$C$ is the number of classes.
$y_{i,c}$ is a binary indicator (0 or 1) that indicates whether class $c$ is the correct label for sample $i$.
$p_{i,c}$ is the predicted probability that sample $i$ belongs to class $c$.

\section*{Contrastive Loss}
The contrastive loss function calculates the loss based on pairs of input samples, typically consisting of a positive pair (similar items) and a negative pair (dissimilar items).

For a pair of samples $x_1$, $x_2$ with a label $y$ indicating whether they are similar (1 for similar, 0 for dissimilar), the contrastive loss is defined as:

\[
L = \frac{1}{2N} \sum_{i=1}^{N} \left( y_i \cdot D^2 + (1 - y_i) \cdot \max(0, m - D)^2 \right)
\]
Where:
$N$ is the number of pairs.
$D$ is the Euclidean distance between the embeddings of the two samples: $D = |f(x_1) - f(x_2)|$.
$m$ is a margin that defines the minimum distance for dissimilar pairs.
$y_i$ indicates if the pair is similar 1 or dissimilar 0.

\section*{Re-Identification (Re-ID) Loss}
Re-Identification (Re-ID) Loss is commonly used in the context of object tracking and image retrieval tasks, particularly to ensure that the same object (e.g., a person or vehicle) is recognized across different frames or images despite changes in appearance, pose, lighting, or background. The goal is to minimize the variations that make identifying the same object difficult.

A typical formulation for Re-ID Loss can involve metric learning approaches, often using the contrastive or triplet loss. Here’s a brief overview of both:

\textbf{Contrastive Loss:}
\[
\text{Loss}_{\text{contrastive}} = (1 - y) \cdot \frac{1}{2} D^2 + (y) \cdot \frac{1}{2} \max(0, m - D)^2
\]
Where:
$D$ is the Euclidean distance between the feature embeddings of two images.
$y$ indicates whether the images belong to the same class (1 if the same, 0 if different).
$m$ is a margin that defines how far apart the positive pairs should be from the negative pairs.

\textbf{Triplet Loss:}
\[
\text{Loss}_{\text{triplet}} = \max(0, D(a, p) - D(a, n) + \alpha)
\]
Where:
$D(a, p)$ is the distance between the anchor $a$ and the positive sample $p$,
$D(a, n)$ is the distance between the anchor $a$ and the negative sample $n$,
$\alpha$ is a margin that ensures that the anchor-positive distance is smaller than the anchor-negative distance by at least $\alpha$.

\section*{Temporal consistency Loss}

Temporal Consistency Loss is a metric used in object tracking to ensure that the tracking predictions remain consistent over time. This loss helps maintain the identity of objects in consecutive frames by minimizing discrepancies in their predicted positions and appearances.

A common way to express Temporal Consistency Loss is through a formulation that compares the current state (e.g., position or features) of an object with its expected state from previous frames. The general formula can be represented as:
\[ L_{temporal} = || f_t - f_{t-1} ||^2 \]
where:
$f_t$ is the feature or state representation of the object at the current time frame $t$,
$f_{t-1}$ is the feature or state representation of the object at the previous time frame $t-1$,
$|| \cdot ||^2$ denotes the $L_2$ norm, measuring the squared Euclidean distance between the two features.

\section*{Cycle consistency Loss}
Cycle consistency loss measures the difference between the original image and the retranslated image after undergoing a cycle of transformations between two image domains. The main idea is that if you translate an image $x$ from domain $X$ to domain $Y$ and then back to domain $X$, the result should ideally be the same as the original image $x$.

For two image domains $X$ and $Y$, where $G$ is the generator that translates from $X$ to $Y$, and $F$ is the generator that translates from $Y$ to $X$, the cycle consistency loss can be defined as:
\[
L_{cycle}(y, G(F(y))) = | y - G(F(y)) |_1
\]
Where:
$x$ is an image from domain $X$.
$y$ is an image from domain $Y$.
$G(x)$ is the image generated by translating $x$ to domain $Y$.
$F(y)$ is the image generated by translating $y$ to domain $X$.
The $| \cdot |_1$ norm typically represents the L1 distance (mean absolute error).
\section*{Pixel-wise mean Loss}
Pixel-wise mean loss is a common loss function used in image processing tasks, particularly in image reconstruction, denoising, dehazing, and segmentation. It measures the difference between the predicted image and the ground truth (actual) image at the level of individual pixels.

The difference between the predicted pixel value and the actual pixel value is computed, usually using a norm. A common choice is the Mean Squared Error (MSE), calculated as follows:
\[
L_{Pixel-wise-mean} = \frac{1}{N} \sum_{i=1}^{N} (P_i - G_i)^2
\]
where $P_i$ is the pixel value from the predicted image, $G_i$ is the pixel value from the ground truth image, and $N$ is the total number of pixels.

\section*{Motion consistency Loss}
Motion consistency loss is a loss function used in computer vision to ensure that the motion represented in frames of a video or in sequential images remains coherent and smooth over time. It encourages models to generate or predict motions that are consistent with the observed data, avoiding abrupt discontinuities.

The motion consistency loss $L_{motion}$ can then be defined as:

\[
L_{motion} = \sum_{(x,y)} \frac{1}{N} \left( \left| \mathbf{u}{pred}(x, y) - 
\mathbf{u}{true}(x, y) \right|^2 + \\ \left| \mathbf{v}{pred}(x, y) - \mathbf{v}{true}(x, y) \right|^2 \right)
\]

Here, $(x,y)$ iterates over all pixel locations, $N$ is a normalization factor (like the number of pixels), $\mathbf{u}{pred}$ and $\mathbf{u}{true}$ are the predicted and true optical flow components.

\section*{SSIM Loss}
SSIM loss, or Structural Similarity Index Measure loss, is a loss function used primarily in image processing tasks, particularly for image reconstruction, super-resolution, and dehazing. Unlike traditional pixel-wise loss functions, such as Mean Squared Error (MSE), which only focus on pixel values, SSIM loss considers structural information in the images, making it more sensitive to perceived visual quality.

SSIM evaluates the similarity between two images based on three main components: Luminance( The brightness of the images), Contrast(The contrast levels of the images), and
Structural Information (The structural layout of the image, which captures patterns and textures).
The SSIM index is calculated using the following formula:
\[
SSIM(x, y) = \frac{(2\mu_x\mu_y + C_1)(2\sigma_{xy} + C_2)}{(\mu_x^2 + \mu_y^2 + C_1)(\sigma_x^2 + \sigma_y^2 + C_2)}
\]
where:

$\mu_x$ and $\mu_y$ are the average pixel values of images $x$ and $y$.
$\sigma_x^2$ and $\sigma_y^2$ are the variances of the pixel values in images $x$ and $y $.
$\sigma_{xy}$ is the covariance between the two images.
$C_1$ and $C_2$ are small constants added to avoid division by zero.

\section*{MS-SSIM Loss}
The Multiscale Structural Similarity Loss (MS-SSIM) is introduced to impose additional constraints on the network, ensuring the accuracy of the reconstructed clear image. This loss function considers the structural similarity at multiple scales, thereby enhancing the perceptual quality of the output. The formula for calculating MS-SSIM involves evaluating the SSIM index at different scales, combining the results to provide a comprehensive measure of similarity between the predicted image and the ground truth. 

The formal definition of the MS-SSIM loss can be represented as:
\[
\text{MS-SSIM}(P, G) = \prod_{i=1}^{n} \left( \frac{(2\mu_{P,i}\mu_{G,i} + C_1)(2\sigma_{PG,i} + C_2)}{(\mu_{P,i}^2 + \mu_{G,i}^2 + C_1)(\sigma_{P,i}^2 + \sigma_{G,i}^2 + C_2)} \right)^{\alpha_i}
\]
where $P$ is the predicted image, $G$ is the ground truth, $\mu$ and $\sigma$ represent the mean and variance for the images, and $C_1$ and $C_2$ are constants to stabilize the division. The parameters $n$ and $\alpha_i$ define the number of scales and the weight for each scale, respectively. By minimizing the MS-SSIM loss, the network can effectively learn to produce high-quality, clear images.

\section*{Negative log-likelihood Loss}
Negative log-likelihood loss evaluates the likelihood of the target data given the predicted probabilities and seeks to minimize this value. It is particularly useful when the output layer uses a softmax activation function to produce probabilities for multi-class classification.

For a single instance, the negative log-likelihood loss is defined as:
\[
L(y, p) = -\log(p_y)
\]
Where:
$y$ is the true class label (one-hot encoded).
$p_y$ is the predicted probability for the class $y$.
For a dataset of $N$ instances, the average negative log-likelihood loss can be expressed as:
\[
L = -\frac{1}{N} \sum_{i=1}^{N} \log(p_{y_i})
\]
Where:
$y_i$ is the true label for the $i-th$ instance.
$p_{y_i}$ is the predicted probability of the true class for the $i-th$ instance.
\section*{Diffusion Loss }
The diffusion loss typically is based on the mean squared error (MSE) between the predicted clean data (or denoised sample) and the actual clean data at each step of the diffusion process. Given a noisy input $x_t$ at time step $t$ and its corresponding clean sample $x_0$, the loss can be defined as:
\[
L_{\text{diffusion}} = \mathbb{E}_{t} \left[ | x_0 - \hat{x}_0(x_t, t) |^2 \right]
\]
Where:
$\hat{x}_0(x_t, t)$ is the model’s prediction of the original sample $x_0$ based on the noisy input $x_t$ at time $t$.
$\mathbb{E}$ represents the expectation over the noise distributions and time steps.


\begin{thebibliography}{1}
\bibitem{yolo5} While not an official release by Joseph Redmon, YOLOv5 has gained traction in the community for its ease of use and performance.
\end{thebibliography}
\end{document}